\pdfoutput=1
\documentclass[12pt]{article}

\usepackage[margin=1.2in]{geometry}
\usepackage{amsmath,amsthm,amssymb}
\usepackage{hyperref}
\usepackage{physics}
\usepackage{enumitem}
\usepackage{graphicx}
\usepackage{subcaption}
\usepackage{cleveref}
\usepackage[svgnames,dvipsnames,x11names,table]{xcolor}
\usepackage{tikz}
\usepackage{pgfplots}
\usetikzlibrary{decorations.markings}
\usetikzlibrary{shapes.misc}
\usetikzlibrary{calc}
\usetikzlibrary{arrows}
\usepackage{subcaption}
\usepackage{cancel}
\usepackage{mathtools}
\usepackage{epigraph}
\usepackage{bm}
\usepackage{tensor}
\usepackage{setspace}
\usepackage{tabularx}

\usepackage[framemethod=default]{mdframed}
\newmdenv[skipabove=7pt,
skipbelow=7pt,
rightline=false,
leftline=false,
topline=false,
bottomline=false,
backgroundcolor=gray!10, 
linecolor=gray,
innerleftmargin=5pt,
innerrightmargin=5pt,
innertopmargin=5pt,
innerbottommargin=5pt,
leftmargin=0cm,
rightmargin=0cm,
linewidth=4pt]{eBox}

\hypersetup{
    colorlinks=true,
    linkcolor={red!50!black},
    citecolor={blue!50!black},
    urlcolor={blue!80!black}
}

\crefname{table}{Table}{Tables}
\crefname{equation}{Eq.}{Eqs.}
\crefname{appendix}{App.}{Apps.}
\crefname{section}{Sec.}{Secs.}
\crefname{figure}{Fig.}{Figs.}
\numberwithin{equation}{section} 

\newcommand{\arr}[1]{{\stackrel{\leftarrow}{#1}}}

\newcommand{\ba}[1]{{\bar{\alpha}_{#1}}}
\newcommand{\FTran}[3]{#1 \left( x_{#2}, t_{#2} | x_{#3}, t_{#3} \right)}
\newcommand{\Tran}[3]{#1 \left( x_{#2} | x_{#3} \right)}
\newcommand{\ITran}[3]{#1 \left( \x_{#2} | \x_{#3} \right)}

\def \d{{\rm d}}

\def \g{{g}}
\def \h{{h}}

\def \D{{\mathbb D}}
\def \F{{\mathbf{F}}} 
\def \H{{H}} 
\def \G{{G}}
\def \I{{K}} 
\def \J{{J}}
\def \K{{\arr{P}}} 

\def \P{{P}} 
\def \Q{{Q}} 
\def \S{{\mathbf{S}}}

\def \s{{s}} 
\def \th{{\bm{\theta}}} 

\def \w{{P}} 
\def \x{{\mathbf{x}}}

\def \E{{\mathbb{E}}} 
\def \F{{\bf F}}

\def \L{{\textrm{L}}}
\def \R{{R}}
\def \T{{T}}

\def \N{{\mathcal N}}
\def \W{{\mathbf{W}}}

\def \pd{{\partial}}

\def \deq {{\,:=\,}}
\def \rdeq {{\,=:\,}}

\def \inP {{\P_{i}}}
\def \finP {{\P_{f}}}
\def \midP {{\w}}
\def \Prob {{\rm Prob}}

\begin{document}

\begin{titlepage}
\setcounter{page}{1} \baselineskip=15.5pt
\thispagestyle{empty}

\begin{center}
{\fontsize{18}{18} \bf Generative Diffusion From An Action Principle}
\end{center}

\vskip 20pt
\begin{center}
\noindent
{\fontsize{12}{18}\selectfont Akhil Premkumar}
\end{center}

\begin{center}
\vskip 4pt
\textit{ {\small Kavli Institute for Cosmological Physics, University of Chicago, IL 60637, USA}
}

\end{center}

\vspace{0.4cm}
 \begin{center}{\bf Abstract}
 \end{center}

\noindent


Generative diffusion models synthesize new samples by reversing a diffusive process that converts a given data set to generic noise. This is accomplished by training a neural network to match the gradient of the log of the probability distribution of a given data set, also called the score. By casting reverse diffusion as an optimal control problem, we show that score matching can be derived from an action principle, like the ones commonly used in physics. We use this insight to demonstrate the connection between different classes of diffusion models.

\end{titlepage}

\setcounter{page}{2}

\restoregeometry

\begin{spacing}{1.2}
\newpage
\setcounter{tocdepth}{2}
\tableofcontents
\end{spacing}

\setstretch{1.1}
\newpage

\section{Introduction}

The field of Generative Artificial Intelligence has witnessed remarkable progress in recent years, fueled by the advent of novel deep learning techniques. Among these advancements, diffusion-based models have emerged as a promising paradigm for generating high-quality, high-dimensional, diverse, and coherent data samples. These models leverage principles from non-equilibrium statistical mechanics to effectively reconstruct the underlying probability distribution from which a training data set was sampled.

The central idea behind diffusion models is \textit{reverse} diffusion. These models gradually add noise to a given data set and observe how the data vectors evolve over time. In a cursory sense, the model learns the forces that were at play in the diffusive stage and uses this information to reverse the transformation and recover the original distribution. By focusing on the forces that furnish the distribution these models sidestep issues of normalizing the probability correctly.

The general problem of transforming a given probability distribution to another was considered by Erwin Schr\"{o}dinger in a seminal paper titled ``On the Reversal of the Laws of Nature" \cite{schrodinger1931,Chetrite:2021gcc}. Schr\"{o}dinger shows that, of all the ways of going from the initial to the final distribution, the most probable transformation is determined by solving a constrained optimization problem. That is, the most likely evolution of the interim probability is the one that extremizes an action which prioritizes the `most efficient' trajectories, whilst satisfying the boundary conditions at either end of the transformation. Reverse diffusion is a specific instance of this problem where the initial distribution is a diffused version of the final one. In that case, the action is fully determined from knowledge of the original diffusion process that we are trying to reverse, specifically the score function. We will show that, under the hood, score matching diffusion models are minimizing this action as they learn to reconstruct data from noise \cite{Sohl-DicksteinW15,Goyal17,Ho2020DDPM,Song2019,Song2021ScoreBased}. There is a close parallel between our approach and the perspectives developed in \cite{Huang21,berner2023optimal}, but ultimately the action principle reproduces the score matching objective more directly.

This work would be most useful to anyone with a math or physics background, who is not necessarily an expert in machine learning or statistics. We only assume a reasonable understanding of differential equations and some calculus of variations on the reader's part. The paper eschews mathematical rigor in favor of physical intuition; we presume sufficient smoothness/slowness for all mathematical functions that appear in our discussions, and avoid explicit use of measure theory. For more formal adaptations of Schr\"{o}dinger's work to diffusion models, the reader is referred to \cite{pavon2022local,Bortoli21,Vargas21,Wang21,Winkler23}.

The paper is structured as follows: \cref{sec:StochasicProcesses} is a review of essential concepts from the theory of stochastic processes. The reader familiar with these ideas can jump ahead to \cref{sec:Schrodinger}, where we recapitulate elements from Schr\"{o}dinger's paper that are foundational to our arguments in subsequent sections. Stochastic optimal control is introduced as a variational problem in \cref{sec:StochastcOptimalControl}. Reverse diffusion is realized as an optimal control problem in \cref{sec:ReversalOptimal}. The corresponding action can be adapted into a score matching objective for diffusion models, as shown in \cref{sec:ScoreMatchingLeastAction}. These considerations are applied to data in \cref{sec:WorkingWithData}, where we explain how reverse diffusion can be implemented with a neural network. A critical step in this process is the conversion of the cost function to a denoising score matching form \cite{Vincent2011}. We examine this closely in \cref{sec:SpecialKernel}, using the kernel from \cite{Sohl-DicksteinW15,Ho2020DDPM}. An intuitive explanation of the score matching is given in \cref{sec:MeaningOfScore}.
There are no experiments attached to this work.



\subsection{Notation}




\begin{table}[h!]
    \centering
    \begin{tabularx}{\textwidth}{@{} p{0.25\textwidth} X @{}}
        \hline
        \textbf{Symbol} & \textbf{Meaning} \\
        \hline
        $x(t), \dot x(t), \ddot x(t)$ & Position, velocity and acceleration of a particle at time $t$. \\
        $F(x,t)$ & Deterministic/drift force acting on the particle. \\
        $D(t)$ & Diffusion coefficient. \\
        $\w(x,t) \d x$ in \cref{sec:StochasicProcesses} & Probability of finding a particle in the region $(x, x+ \d x)$ at time $t$. That is, $P(x,t)$ is a probability density function. \\
        $\P(x,t|x_0,t_0) \d x$ & The probability of finding a particle in $(x, x+ \d x)$ at time $t$, given that it started out at $x_0$ at time $t_0$. Also called the \textit{transition probability}. \\
        $\Tran{P}{\s+1}{\s}$ & The transition probability over a small time step $\Delta t_s$. Equivalent to $\P(x_{\s+1},t_{\s+1}|x_{\s},t_{\s})$, and used for brevity. \\
        $\N(x;\mu,\sigma^2)$ & A normalized Gaussian distribution in $x$, with mean $\mu$ and variance $\sigma^2$. $x \sim \N(\mu,\sigma^2)$ is a sample of this distribution. \\
        $W(t)$ & The Wiener process. \\
        ${\rm D}_{KL}$ & The Kullback-Leibler divergence. \\
        $P(x,t)$ in \cref{sec:TransformingProbs}, \ref{sec:Unification} & Intermediate probability during the process $\inP \to \finP$. \\
        $\x$ & A data vector in $d$-dimensional space. \\
        $\lVert \x \rVert^2_2$ & The $\ell^2$ norm of the vector $\x$. \\
        $\th$ & Weights of a neural network. \\
        $\mathbb{E}_{\x}[f(\x)]$ & The expectation value of a function $f(\x)$. Equivalent to $\frac{1}{N} \sum_{\x} f(\x) \asymp  \int \d \x P(\x) f(\x)$, for large number of samples $N$. \\
        $O(\delta^n)$ & A term that is order $\delta^n$ in magnitude. \\
        l.h.s./r.h.s. & left hand side/right hand side. \\
        \hline
    \end{tabularx}
    \caption{\label{tab:Notation} Symbols used in this work and their meaning.}
\end{table}


\newpage

\section{Stochastic Processes}
\label{sec:StochasicProcesses}

We begin with a brief review of stochastic processes, focusing on the concepts required for sections that follow.

\subsection{The Langevin and Fokker-Planck Equations}

Consider a particle moving through space under the combined influence of a deterministic force $F$, and random noise $\eta$. The motion of the particle is governed by the equation
%
\begin{equation}
    m \ddot x(t) + \gamma \dot x(t) = F(x,t) + \eta(t) ,
\end{equation}
%
where $m$ is the mass of the particle, and $\gamma$ is the coefficient of drag.
If the system is dominated by friction we can discard the acceleration term, since $\gamma \dot x \gg m \ddot x$. A typical example is Brownian motion of a free particle in a very viscous fluid. We may then choose units such that $\gamma = 1$, arriving at the \textit{Langevin equation}
\begin{equation}
    \dot x(t) = F(x,t) + \eta(t) . \label{eq:Langevin}
\end{equation}
Since this equation involves only a first derivative in time, it describes a Markovian (memoryless) process. We will work with Gaussian noise throughout this paper, which means $\eta$ is completely characterized by its one and two point correlation functions
\begin{equation}
    \langle \eta(t) \rangle = 0, \qquad
    \langle \eta(t) \eta(\tau) \rangle = D(t) \delta (t-\tau) .
    \label{eq:NoiseCorrelations}
\end{equation}
The quantity $D(t)$ is called the diffusion coefficient, and it is a positive definite number that specifies the strength of the noise. In the Brownian example, the noise is generated by random collisions of the particle with the molecules of the liquid. These collisions are random in time, direction, and strength, and are well approximated by \cref{eq:NoiseCorrelations} \cite{einstein1956investigations}. \cref{eq:Langevin} admits infinitely many solutions, each of which describes some particular trajectory of the particle over time.

An alternative, but equivalent, approach to describe the same dynamics is to study the ensemble behavior of a large number of such particles subject to the same $F$ and $\eta$. Suppose we release several particles from a position $x_0$ at time $t_0$, and observe their motion under \cref{eq:Langevin}. At some time $t>t_0$ the particles would be distributed across space, and we can assign a probability $\P(x, t | x_0, t_0) \d x$ of finding a particle in a region $(x, x+ \d x)$. The time evolution of the probability density function  $\P(x, t | x_0, t_0)$ is governed by the \textit{Fokker-Planck} equation\footnote{A derivation of the Fokker-Planck equation from \cref{eq:Langevin} is given in chapter 4 of \cite{gardiner2004handbook,Zinn-Justin:2002}.}
\begin{equation}
    \pd_t \P(x, t | x_0, t_0)
        = - \pd_x \left( F(x,t) \P(x, t | x_0, t_0) \right)
            + \frac{1}{2} D(t) \pd_x^2 \P(x, t | x_0, t_0) .
    \label{eq:FokkerPlanckTransition}
\end{equation}
The object $\P(x, t | x_0, t_0)$ is a transition probability, or a Green's function, since it can be used to evolve an arbitrary initial distribution of the particles, $\P_0(x_0)$, to the present time $t$:
\begin{equation}
    \w(x, t) = \int_{-\infty}^{\infty} \d x_0 \P(x, t | x_0, t_0) \w_0(x_0) .
    \label{eq:GreensFn}
\end{equation}
In other words, \cref{eq:GreensFn} is the solution to
\begin{equation}
    \pd_t \w(x,t)
        = - \pd_x \left( F(x,t) \w(x, t) \right)
            + \frac{1}{2} D(t) \pd_x^2 \w(x,t) ,
    \label{eq:FokkerPlanck}
\end{equation}
with the initial condition $\P(x_0, t_0) = \w_0(x_0)$.

\subsection{Formal Solution}
\label{sec:FormalSolution}

We can gain insight into the behavior of the random trajectories generated from \cref{eq:Langevin} by discretizing the time variable. We will assume that $F(x,t)$ and $D(t)$ are smooth, deterministic functions, and forego any discussion about the existence and convergence of the discretized expressions in the continuum limit (such details are covered in Chapter 5 of \cite{schuss2010}). To begin, we consider \cref{eq:Langevin,eq:FokkerPlanck} with the drift term set to zero,
\begin{subequations}
    \begin{align}
        \dot x &= \eta(t) , \label{eq:LangevinDiffusion} \\
        \pd_t \w(x,t) &= \frac{D(t)}{2} \pd_x^2 \w(x,t) . \label{eq:FPDiffusion}
    \end{align}
\end{subequations}
This describes a purely diffusive process. If $D(t)$ is set to a constant $D$, \cref{eq:FPDiffusion} is solved by the kernel (cf.\ \cref{eq:GreensFn})
\begin{align}
    \P(x,t|x_0,t_0)
        &= \frac{1}{\sqrt{2 \pi D (t-t_0)}} \exp[- \frac{(x-x_0)^2}{2 D (t-t_0)}] \\[0.5em]
        &\equiv \N(x; x_0, D(t-t_0)) , \label{eq:PureDiffusionKernel}
\end{align}
where $\N(x, \mu, \sigma^2)$ denotes a Gaussian distribution with mean $\mu$ and variance $\sigma^2$. In this particular case, the Markov property \cref{eq:MarkovProperty} follows from the fact that the convolution of two Gaussians in another Gaussian,
\begin{equation}
    \int_{-\infty}^{\infty} \d \xi \N(\xi; \mu_1, \sigma_1^2) \N(x-\xi; \mu_2, \sigma_2^2)
    = \N(x; \mu_1 + \mu_2, \sigma_1^2 + \sigma_2^2) . \label{eq:ConvolutionOfGaussians}
\end{equation}
Notice how the means and variances simply add up under this operation. This fact can be used to compute the transition probability for the case of generic $D(t)$. We introduce $n+1$ points $t_0, t_1, t_2 \dots, t_n$ in the interval $[t_0, t]$, with $t = t_n$ and $t_{\s+1} - t_{\s} = \Delta t_{\s}$. If $\Delta t_{\s}$ is small enough $D(t_{\s})$ is nearly constant\footnote{If we had $D(x)$ rather than $D(t)$, we would need to be more careful about where the particle was when it gets a random kick at time $t$. For example, we could use the strength of the kick evaluated at $x_{\s}$ (It\^{o}), or model the kick as a force that is applied over a small but finite time, so that the average strength $[D(x_{\s+1})+D(x_{\s})]/2$ must be used (Stratonovich).} between $t_{\s}$ and $t_{\s+1}$. Denoting the position of the particle at $t_{\s}$ as $x_{\s}$, we may then write
%
%
\begin{subequations}
    \begin{gather}
        \P(x,t|x_0,t_0) =
            \int \prod_{\s=1}^{n-1} \d x_{\s}
            \Tran{\P}{\s+1}{\s} \Tran{\P}{1}{0} \label{eq:TrotterizedTransition} \\
        \Tran{\P}{\s+1}{\s} = \frac{1}{\sqrt{2 \pi D(t_{\s}) \Delta t_{\s}}} \exp[- \frac{(x_{\s+1}-x_{\s})^2}{2 D(t_{\s}) \Delta t_{\s}}] . \label{eq:SmallTransitionDiffusion}
    \end{gather}
    \label{eq:Trotterized}
\end{subequations}
In the limit $n \to \infty, \Delta t_{\s} \to 0$, with $\sum_{\s=0}^{n-1} \Delta t_{\s} = t-t_0$ kept fixed, \cref{eq:TrotterizedTransition} is the formal solution to \cref{eq:FPDiffusion}, called the \textit{path integral} \cite{Zinn-Justin:2002}.
The path integral accumulates the probabilities of all paths that take the particles from $(x_0,t_0)$ to $(x,t)$. 

Intuitively, \cref{eq:Trotterized} is an instruction to choose $x_{\s+1}-x_{\s}$ from a Gaussian with mean zero and variance $D(t_{\s}) \Delta t_{\s}$ at each time step. That is,
%
\begin{equation}
    x_{\s+1} - x_{\s} \sim \N(0, D(t_{\s}) \Delta t_{\s}) . \label{eq:JumpGeneric1}
\end{equation}
The special case of $D(t)=1$ is called a \textit{Wiener process}, and it is conventional to rename $x(t) \to W(t)$ to indicate this. Then,
\begin{equation}
    \Delta W_\s \deq W_{\s+1} - W_\s \sim \N(0, \Delta t_{\s}), 
    \label{eq:WienerProcess}
\end{equation}
in terms of which we can write \cref{eq:JumpGeneric1} as
\begin{equation}
    x_{\s+1} - x_{\s} = \sqrt{D(t_{\s})} \Delta W(t_{\s}) , \label{eq:JumpGeneric2}
\end{equation}
This is the physical content of the stochastic differential equation (SDE)\footnote{The object $\d W(t)$ has a deeper meaning in stochastic calculus. The interested reader may refer chapter 4 of \cite{gardiner2004handbook} for more details.}
\begin{equation}
    \d x(t) = \sqrt{D(t)} \, \d W(t).
\end{equation}

We can repeat these arguments to find the transition probability when a drift term present,
\begin{subequations}
    \begin{align}
        \dot x &= F(x,t) + \eta(t) , \\
        \pd_t \w(x,t) &= -\pd_x(F(x,t) \w(x,t)) + \frac{D(t)}{2} \pd_x^2 \w(x,t) .
    \end{align}
    \label{eq:WithDrift}
\end{subequations}
In a small time step $\Delta t_{\s}$ the particle is displaced by $F(x_{\s}, t_{\s}) \Delta t_{\s}$ in addition to the random kick from \cref{eq:JumpGeneric2},
\begin{equation}
    x_{\s+1} - x_{\s} = F(x_{\s}, t_{\s}) \Delta t_{\s} + \sqrt{D(t_{\s})} \Delta W(t_{\s}) . \label{eq:JumpWDrift}
\end{equation}
Therefore $\P(x,t|x_0,t_0)$ still has the same form as \cref{eq:TrotterizedTransition}, but the transition probability in each time step has to be modified to
\begin{equation}
    \Tran{\P}{\s+1}{\s} =
        \frac{1}{\sqrt{2 \pi D(t_{\s}) \Delta t_{\s}}}
        \exp[- \frac{\Delta t_{\s}}{2 D(t_{\s})} \left( \frac{x_{\s+1}-x_{\s}}{\Delta t_{\s}} - F(x_{\s}, t_{\s}) \right)^2 ] .
    \label{eq:SmallTransitionWDrift}
\end{equation}
Crucially, this is a Gaussian in the variable $x_{\s+1}$ regardless of the functional form of the drift force $F(x,t)$.
Finally, we note that \cref{eq:JumpWDrift} corresponds to the SDE
\begin{equation}
    \d x(t) = F(x,t) \d t + \sqrt{D(t)} \, \d W(t) .
    \label{eq:LangevinSDE}
\end{equation}

\subsection{Kolmogorov Equations}
\label{sec:ForwardAndBackward}

The Fokker-Planck equation, \cref{eq:FokkerPlanckTransition}, is also called the \textit{forward} Kolmogorov equation, as it involves differential operators with respect to $x$ and $t$, the coordinates of the particle at the later time $t > t_0$. The corresponding backward equation is an equation of motion for $\P(x,t|x_0,t_0)$ where we differentiate with respect to $x_0$ and $t_0$, the particle coordinates at the earlier time. To derive the backward equation we start with the Chapman-Kolmogorov equation,
\begin{equation}
    \int_{-\infty}^{\infty} \d \xi \P(x, t | \xi, \tau) \P(\xi, \tau | x_0, t_0) = \P(x, t | x_0, t_0) , \label{eq:MarkovProperty}
\end{equation}
where $\tau \in (t_0, t)$ is some intermediate time, and $\xi$ is the position of the particle at that instant. Differentiating both sides with respect to $\tau$,
\begin{gather}
    \frac{\pd}{\pd \tau} \int_{-\infty}^{\infty} \d \xi \P(x, t | \xi, \tau) \P(\xi, \tau | x_0, t_0) = 0 \\ \implies
    \int_{-\infty}^{\infty} \d \xi \left[ \pd_{\tau} \P(x, t | \xi, \tau) \right] \P(\xi, \tau | x_0, t_0) + \P(x, t | \xi, \tau) \left[ \pd_{\tau} \P(\xi, \tau | x_0, t_0) \right] = 0 . \label{eq:BackwardFPStep1}
\end{gather}
We can use \cref{eq:FokkerPlanckTransition} on the second term in the integrand, and integrate by parts to transfer the $\pd_{\xi}$ operators to $\P(x, t | \xi, \tau)$. The boundary terms are assumed to vanish at $\xi \to \pm \infty$. This leaves us with
\begin{equation}
    \int \d \xi \left[ \left( \pd_{\tau} + F(\xi, \tau) \pd_{\xi} + \frac{1}{2} D(\tau) \pd_{\xi}^2 \right) \P(x, t | \xi, \tau) \right] \P(\xi, \tau | x_0, t_0) = 0 .
\end{equation}
Since this relation must hold for any $x$ and $t$, the integrand must vanish everywhere. Thus, we arrive at the \textit{backward} Kolmogorov equation
\begin{equation}
    \pd_{\tau} \P(x, t | \xi, \tau) = - F(\xi, \tau) \pd_{\xi} \P(x, t | \xi, \tau) - \frac{1}{2} D(\tau) \pd_{\xi}^2 \P(x, t | \xi, \tau) . \label{eq:BackwardFP1}
\end{equation}
We collect the forward and backward equations below for convenience and express them in terms of the same variables (with $t > \tau$). Once again, we point out that both these equations are solved by the \textit{same} function $\P(x, t | \xi, \tau)$:
\begin{equation}
    \begin{aligned}
        \pd_{t} \P(x, t | \xi, \tau) &= - \pd_{x}(F(x,t) \P(x, t | \xi, \tau)) + \frac{1}{2} D(t) \pd_{x}^2 \P(x, t | \xi, \tau) \rdeq \L^\dagger_x \P(x, t | \xi, \tau) \\
        \pd_{\tau} \P(x, t | \xi, \tau) &= - F(\xi, \tau) \pd_{\xi} \P(x, t | \xi, \tau) - \frac{1}{2} D(\tau) \pd_{\xi}^2 \P(x, t | \xi, \tau) \rdeq -\L_\xi \P(x, t | \xi, \tau).
    \end{aligned}
    \label{eq:ForwardAndBackwardFP}
\end{equation}
We know from \cref{eq:GreensFn} that $\P(x, t | \xi, \tau)$ evolves a given initial distribution forward in time, as \cref{eq:FokkerPlanck}. In a similar manner, the \textit{terminal value problem}
\begin{subequations}
    \begin{gather}
        \pd_{\tau} \J(\xi, \tau) = - F(\xi, \tau) \pd_{\xi} \J(\xi, \tau) - \frac{1}{2} D(\tau) \pd_{\xi}^2 \J(\xi, \tau) 
        \label{eq:BackwardKolmogorov} \\
        \lim_{\tau \to t} \J(\xi, \tau) = \J_1(\xi) , \label{eq:Terminal}
    \end{gather}
\end{subequations}
has the solution
\begin{equation}
    \J(\xi, \tau)
    = \int_{-\infty}^{\infty} \d x \J_1(x) \P(x, t | \xi, \tau)
    \rdeq \E_{\rm \P} \left[ \J_1(x(t)) \big| x(\tau) = \xi \right] .
    \label{eq:GreenFnBackward}
\end{equation}
This expression can be understood as follows: $\P(x, t | \xi, \tau)$, seen as a path integral, adds up the probabilities of all trajectories from $(\xi,\tau) \to (x,t)$ under the evolution in \cref{eq:LangevinSDE} (we call this process $\textrm{\P}$). We start with a large number of particles at $(\xi, \xi + \d \xi)$ at time $\tau$, and insist that fraction $\J_1(x) \d x$ of particles land on $(x, x+ \d x)$ at a later time $t$. This can be arranged by taking a weighted sum of the path integrals $\P(x, t | \xi, \tau)$, weighed by $\J_1(x)$. Therefore, $\J(\xi, \tau)$ is the probability that particles released at $(\xi,\tau)$, and subject to $\textrm{\P}$, are distributed as $\J_1(x)$ at $t$. The meaning of the conditional expectation introduced above should now be clear.



\subsection{Killed Diffusion}

The backward Kolmogorov equation can be generalized by adding an extra term,
\begin{equation}
    \pd_{\tau} \J(\xi, \tau) + F(\xi, \tau) \pd_{\xi} \J(\xi, \tau) + \frac{1}{2} D(\tau) \pd_{\xi}^2 \J(\xi, \tau) - V(\xi, \tau) \J(\xi, \tau) = 0 .
    \label{eq:BackwardGeneralized}
\end{equation}
The solution to \cref{eq:BackwardGeneralized}, with the terminal condition \cref{eq:Terminal}, is given by the \textit{Feynman-Kac formula}:
\begin{align}
    \J(\xi, \tau)
    &= \E_{\rm \P} \left[ \J_1(x(t)) \exp(-\int_{\tau}^{t} \d \bar{t} \, V(\bar{x}, \bar{t})) \bigg| x(\tau) = \xi \right] \\[0.5em]
    &= \int_{-\infty}^{\infty} \d x \J_1(x) \I(x, t | \xi, \tau) ,
    \label{eq:FeynmanKac}
\end{align}
%
where $\I(x,t|\xi,\tau)$ is a new kernel defined as
\begin{equation}
    \I(x,t|\xi,\tau) \deq \P(x,t|\xi,\tau) \exp(-\int_{\tau}^{t} \d \bar{t} \, V(\bar{x}, \bar{t})) .
    \label{eq:KilledKernel}
\end{equation}
The exponential factor can be absorbed into the path integral representation \cref{eq:TrotterizedTransition} by writing $\int \d \bar{t} \, V(\bar{x}, \bar{t}))$ as $\sum_{\s} \Delta t_{\s} V(x_{\s}, t_{\s})$. This leads us to the following transition probability over a small time interval:
\begin{equation}
    \Tran{\I}{\s+1}{\s} =
        \frac{1}{\sqrt{2 \pi D(t_{\s}) \Delta t_{\s}}}
        \exp[- \frac{\left( x_{\s+1}-x_{\s} - F(x_{\s}, t_{\s}) \Delta t_{\s} \right)^2}{2 D(t_{\s}) \Delta t_{\s}} - V(x_{\s}, t_{\s}) \Delta t_{\s} ] .
        \label{eq:SmallTransitionWKilling}
\end{equation}
The additional term in \cref{eq:BackwardGeneralized} has a simple interpretation: the number of particles that make the transition $x_{\s} \to x_{\s+1}$ is augmented by a factor $ \approx 1 - V(x_{\s}, t_{\s}) \Delta t_{\s}$ at each time step. If $V < 0$, a small fraction of the particles are added to the process during $\Delta t_s$, whereas $V > 0$ implies some particle trajectories terminate in that interval. Therefore $V$ is referred to as the \textit{killing rate}, and \cref{eq:BackwardGeneralized} is a \textit{killed diffusion} process.
It follows from \cref{eq:KilledKernel} that $\I(x,t|\xi,\tau)$ satisfies
\begin{equation}
    \begin{aligned}
        \pd_{t} \I(x, t | \xi, \tau) &= - \pd_{x}(F(x,t) \I(x, t | \xi, \tau)) + \frac{1}{2} D(t) \pd_{x}^2 \I(x, t | \xi, \tau) - V(x, t) \I(x, t | \xi, \tau) \\
        \pd_{\tau} \I(x, t | \xi, \tau) &= - F(\xi, \tau) \pd_{\xi} \I(x, t | \xi, \tau) - \frac{1}{2} D(\tau) \pd_{\xi}^2 \I(x, t | \xi, \tau) + V(\xi, \tau) \I(x, t | \xi, \tau) .
    \end{aligned}
    \label{eq:ForwardAndBackwardKilling}
\end{equation}

\subsection{The Ornstein-Uhlenbeck Process}
\label{sec:OUprocess}


Consider a particle subject to an affine drift force $F = - x$. If the particle is displaced from $x = 0$, it experiences a restoring force that pulls it back toward that point, the strength of the force growing larger as the particle moves farther away. On the other hand, diffusion tends to spread the particles around, setting up a competition between $F$ and $\eta$. This is an example of an \textit{Ornstein-Uhlenbeck process} \cite{Ornstein1930}. The more general case is defined by the SDE
%
\begin{equation}
    \d x(t) = -\frac{\beta(t)}{2} x \, \d t + \sqrt{D(t)} \, \d W(t) .
    \label{eq:OUSDE}
\end{equation}
The finite time transition probability for this process has the closed form solution \cite{risken1989fpe}
%
\begin{equation}
    \P(x, t | x_0, t_0) = \sqrt{\frac{\beta(t)}{2 \pi D(t)} \cdot \frac{1}{ 1 - e^{-\int_{t_0}^t \beta (\bar t) \d \bar t}}}
    \exp[ -\frac{\beta(t)}{2 D(t)} \cdot \frac{\left(x - x_0 e^{-\frac{1}{2} \int_{t_0}^t \beta (\bar t) \d \bar t} \right)^2}{ 1 - e^{-\int_{t_0}^t \beta (\bar t) \d \bar t}} ] .
    \label{eq:OUkernel}
\end{equation}
If $\beta(t)$ and $D(t)$ eventually reach steady state values (say $\beta$ and $D$), so will $\w(x, t)$ (cf.\ \cref{eq:GreensFn}):
\begin{equation}
    \P(x, t \to \infty) = \sqrt{\frac{\beta}{2 \pi D}} e^{-\beta x^2/2 D} . \label{eq:AsymptoticOU}
\end{equation}
That is, \textit{any} initial probability distribution evolves to a stationary Gaussian under \cref{eq:OUSDE}, with appropriate parameters. A special version of \cref{eq:OUSDE} will be discussed in \cref{sec:DDPM}.

\section{Transforming Probabilities}
\label{sec:TransformingProbs}

We are interested in the general problem of morphing a given distribution $\inP(x)$ to a target distribution $\finP(x)$. Reverse diffusion is a particular instance of this problem wherein $\inP(x)$ is a diffused version of $\finP(x)$. To build some intuition for the general case we consider the following \textit{gedanken} experiment.

Imagine placing a drop of ink into a beaker containing still water. We then turn our attention away from the beaker for a while. During this time Brownian motion sets in, and we expect the ink particles to diffuse into the liquid, distributing the ink homogeneously throughout the beaker. Suppose we return to our experiment and find that the drop of ink did \textit{not} diffuse in this manner, but instead, collected into a fairly localized cloud at some corner of the beaker. This is highly surprising.

Puzzled by this outcome we repeat the experiment several more times, with identical lab equipment, and find that the ink particles homogenized in the expected way in all subsequent experiments. So there were no unknown influences at play in our first trial; we just happened to observe an extremely unlikely outcome of our experiment.

One might wonder if such an outcome is allowed at all. With proper normalization, we can interpret the density distribution $\rho(x,t)$ of the ink particles as a probability distribution $P(x,t)$. Then, the final distribution of the ink particles \textit{must} be given by \cref{eq:GreensFn}. In other words, if $\rho(x,t)$ evolves under the diffusion equation $\pd_t \rho = \frac{D}{2} \pd_x^2  \rho$, the boundary condition at $\rho(x,t_i) = \rho_i(x)$ would completely specify the density distribution for all $t > t_i$. In light of this, how is a localized cloud configuration even possible?

In \cref{sec:FormalSolution} we partitioned the time interval $[t_0,t]$ into many smaller ones, and argued that at each time step the particle's jumps are sampled from a Gaussian, \cref{eq:SmallTransitionDiffusion}. Since that distribution is non-zero for all jumps, and the particle trajectory is a sequence of such jumps, it must be that all possible trajectories are allowed by diffusion. The paths that take each ink particle from the original drop and assemble them as a localized cloud are therefore possible, albeit improbable. What then, is the probability of observing a given ink configuration at the end of our experiment? The answer to this question was given by Schr\"{o}dinger in 1931 \cite{schrodinger1931,Chetrite:2021gcc}.

\subsection{Schr\"{o}dinger's Argument}
\label{sec:Schrodinger}

We start our experiment by releasing a large number of particles, $N$, at time $t_i$. We divide the $x$-axis into cells of unit length and denote by $a_l$ the number of particles that start from the $l^{\rm th}$ cell at $t_i$. At the end of our analysis we will take
\begin{equation}
    a_l \to N \inP(x_i) \d x_i , \label{eq:InitialParticleDistribution}
\end{equation}
where the $l^{\rm th}$ cell covers the segment $(x_i, x_i + \d x_i)$, and $\inP(x_i)$ can be interpreted as the initial probability distribution. The particles are allowed to diffuse freely till time $t_f$, at which point $b_k$ particles end up in the $k^{\rm th}$ cell. In the continuum limit this cell covers $(x_f, x_f + \d x_f)$, so that
\begin{equation}
    b_k \to N \finP(x_f) \d x_f .
\end{equation}
Importantly, $\finP(x_f)$ is not the expected outcome of diffusion,\footnote{\label{fn:FandDvalue} To be specific, we can pick a process with $F(x,t) = 0$ and $D(t) = D$ here. So the transition probability $\G(x_f,t_f|x_i,t_i)$ is the function from \cref{eq:PureDiffusionKernel}. In subsequent sections $\G$ will refer to more general Fokker-Planck kernels. See \cref{tab:Notation}.}
\begin{equation}
    \finP(x_f) \neq \int_{-\infty}^{\infty} \d x_i \G(x_f,t_f|x_i,t_i) \inP(x_i) ,
    \label{eq:UnexpectedOutcome}
\end{equation}
We denote the discretized version of $\G(x_f,t_f|x_i,t_i)$ by $\g_{kl}$. That is, $\g_{kl}$ gives the probability that a particle which starts at the $l^{\rm th}$ cell at $t_i$ makes its way to the $k^{\rm th}$ cell at $t_f$. Finally, let $c_{kl}$ be the number of particles that arrive at the $k^{\rm th}$ cell from the $l^{\rm th}$. These $c_{kl}$ must satisfy
\begin{equation}
    \begin{aligned}
        \sum_{k} c_{kl} &= a_l \quad \text{for any } l, \\
        \sum_{l} c_{kl} &= b_k \quad \text{ for any } k .
    \end{aligned}
    \label{eq:BoundaryConditions1}
\end{equation}
Furthermore, since the total number of particles is conserved,
\begin{equation}
    \sum_{k} a_l = \sum_{l} b_k = N .
\end{equation}
Although $a_l$ and $b_k$ are fixed, the conditions \cref{eq:BoundaryConditions1} can be realized by many different matrices $c_{kl}$, depending on how many particles we transfer between the cells. In turn, each choice of $c_{kl}$ corresponds to a family of trajectories, and the optimal $c_{kl}$ will be the one that has the most probable trajectories, consistent with diffusion.

The probability that $c_{kl}$ particles are transported from the $l^{\rm th}$ cell to the $k^{\rm th}$ one is $\g_{kl}^{c_{kl}}$. Since the $l^{\rm th}$ cell has $a_l$ particles, we must account for the number of ways we can choose $c_{kl}$ particles from these. Therefore, the particles in the $l^{\rm th}$ cell migrate to the $k^{\rm th}$ cells with probability
\begin{equation}
    \prod_{k} \frac{a_l!}{c_{kl}!} \g_{kl}^{c_{kl}} ,
\end{equation}
For a given $c_{kl}$, the overall probability for migrations from all such $l$ is therefore
\begin{equation}
    \Prob(c_{kl}) = \prod_{l} a_l! \prod_{k} \frac{\g_{kl}^{c_{kl}}}{c_{kl}!}
\end{equation}
In the large $N$ limit, we can use Stirling's approximation ($\ln n! \approx n \ln n - n$) to write
\begin{equation}
    \Prob(c_{kl}) \asymp \exp[ \sum_{kl} \left( -c_{kl} \ln{\frac{c_{kl}}{\g_{kl}}} + c_{kl} \right) + \sum_l \left( a_l \ln a_l - a_l \right) ] .
    \label{eq:ProbCkl}
\end{equation}
To make the dependence on $N$ explicit, we can express $a_l$ and $c_{kl}$ as
\begin{subequations}
    \begin{align}
        a_l &= N p_i(l) \\
        c_{kl} &= \h(k|l) a_l = N \h(k|l) p_i(l) ,
    \end{align}
    \label{eq:NumbersAsDistr}
\end{subequations}
where $p_i(l)$ (or $p_f(k)$) is just the initial distribution $P_i(x)$ (or final distribution $P_n(x)$), appropriately discretized (cf.\ \cref{eq:InitialParticleDistribution}). Then, $\h(k|l)$ is the empirical transition probability from $p_i(l)$ to the final distribution $p_f(k)$. From \cref{eq:BoundaryConditions1} and \cref{eq:NumbersAsDistr},
\begin{equation}
    b_k = N \sum_k \h(k|l) p_i(l) = N p_f(k) .
    \label{eq:hPropagator}
\end{equation}
Substituting \cref{eq:NumbersAsDistr} into \cref{eq:ProbCkl} and simplifying, we arrive at the formula
\begin{equation}
    \Prob(c_{kl}) \asymp \exp[ -N \sum_{kl} p_i(l) \h(k|l) \ln{\frac{\h(k|l)}{\g(k|l)}} ] ,
\end{equation}
where we have adjusted the notation a bit, writing $\g(k|l) \equiv \g_{kl}$ to highlight the fact that the term in the exponent is comparing two transition probabilities. This term is readily identifiable as the \textit{Kullback-Leibler divergence} between these distributions, averaged over the the initial probability $p_i(l)$,
\begin{equation}
    {\rm D}_{KL}({\rm \h || \g}) \deq \sum_{kl} p_i(l) \h(k|l) \ln{\frac{\h(k|l)}{\g(k|l)}} ,
    \label{eq:DKLDiscrete}
\end{equation}
so that
\begin{equation}
    \Prob(c_{kl}) \asymp e^{ -N {\rm D}_{KL}({\rm \h || \g}) } . \label{eq:ProbCklAsymp}
\end{equation}
It is clear that any outcome satisfying \cref{eq:UnexpectedOutcome} is highly improbable if there are a large number of particles in our experiment; the largeness of $N$ and the convexity of the Kullback-Leibler divergence punish us for deviating from the expected result (cf.\ \cref{eq:DKLPositivity}). Therefore, of all the $c_{kl}$ consistent with \cref{eq:BoundaryConditions1}, the $c_{kl}^\star$ for which the Kullback-Leibler divergence is minimum will dominate the probability of arriving at the configuration $\{b_k\}$ starting from $\{a_l\}$,
\begin{equation}
    \Prob \left( \{b_k\} | \{a_l\} \right) = \sum_{\{c_{kl}\}} \Prob(c_{kl})
    \overset{N \to \infty}{\approx} \Prob(c_{kl}^\star) .
    \label{eq:ProbOverwhelming}
\end{equation}
Of the multitude of trajectories that take $a_l \to b_k$, only the most probable ones contribute to $c_{kl}^\star$, with the rest being exponentially suppressed in the limit of large $N$. This is an example of the \textit{Large Deviation Principle}.\footnote{Whenever we compute an entropy (like \cref{eq:DKLDiscrete}) or a free energy, large deviation theory is at work. See Sec.\ 2 of \cite{Cohen:2022clv} for a quick review, or \cite{Touchette_2009} for a more extensive one.}
%

The question we started with has now been answered: to find the probability of getting an unexpected ink distribution, enumerate all empirical transition probabilities $\h(k|l)$ that take us from the initial configuration to the final one. Out of these, choose the one that is closest to the inherent transition probability $\g(k|l)$, in the Kullback-Leibler sense. We denote this choice by $\h(k|l)^\star$. The required probability is just \cref{eq:ProbCklAsymp} evaluated at $\h(k|l)^\star$.

\subsection{Stochastic Optimal Control}
\label{sec:StochastcOptimalControl}

We are now in a position to address the problem of transforming $\inP \to \finP$ under the process $\textrm{G}$. In particular, what is the probability $\midP(x,t)$ at an intermediate time $t \in (t_i,t_f)$, that specifies the most likely evolution between $\inP$ and $\finP$? With a few technical refinements, this question can be formulated as a variational problem for an action functional. The latter will lead us to the training objective for generative diffusion models in \cref{sec:ScoreMatchingLeastAction}.

We begin by translating \cref{eq:DKLDiscrete,eq:ProbCklAsymp} back to continuum variables:
\begin{gather}
    {\rm D}_{KL}({\rm \H || \G})
        \deq \int \d x_f \d x_i \inP(x_i) \FTran{\H}{f}{i} \ln \frac{\FTran{\H}{f}{i}}{\FTran{\G}{f}{i}} , \label{eq:DKLContinuum} \\[0.5em]
    \Prob(\textrm{\H} ; \inP ) \asymp e^{ -N {\rm D}_{KL}({\rm \H || \G}) } . \label{eq:ProbHAsymp}
\end{gather}
In order to compute $\midP(x,t)$ we must find the empirical transition probability $\H^\star$ that minimizes \cref{eq:DKLContinuum}, whilst also satisfying the continuum version of \cref{eq:hPropagator},
%
\begin{equation}
    \finP(x_f) = \int_{-\infty}^{\infty} \d x_i \FTran{\H}{f}{i} \inP(x_i) .
    \label{eq:HPropagator}
\end{equation}
This is the problem of \textit{optimal control}. To simplify \cref{eq:DKLContinuum} further, we reconstitute it in terms of incremental transitions over many small time intervals, using \cref{eq:TrotterizedTransition}. This is done in \cref{sec:KLDivergence} and results in the inequality
\begin{equation}
    {\rm D}_{KL}({\rm \H || \G})
        \leq \sum_{\s=0}^{n-1} \int \d x_{\s+1} \d x_{\s} \, \midP(x_{\s},t_{\s}) \Tran{\H}{\s+1}{\s} \ln \frac{\Tran{\H}{\s+1}{\s}}{\Tran{\G}{\s+1}{\s}} .
    \label{eq:DKLBound}
\end{equation}
The object on the right, called the pathwise Kullback-Leibler divergence, can be evaluated explicitly for kernels of the form in \cref{eq:SmallTransitionWDrift,eq:SmallTransitionWKilling}. Specifically, let $\textrm{H}$ be a Fokker-Planck process with drift $F_{\textrm{H}}$, and $\textrm{G}$ be a killed diffusion process with drift $F_{\textrm{G}}$ and killing rate $V_{\textrm{G}}$. Both processes have the same diffusion coefficient $D(t)$. Then,
\begin{equation}
    {\rm D}_{KL}({\rm \H || \G}) \leq
        \sum_{\s=0}^{n-1}
        \Delta t_{\s} \int_{-\infty}^{\infty} \d x_{\s} \midP(x_{\s},t_{\s}) \left( \frac{\left| F_{\textrm{\H}}(x_{\s},t_{\s}) - F_{\textrm{\G}}(x_{\s},t_{\s}) \right|^2}{2 D(t_{\s})} + V_{\textrm{G}}(x_s,t_s) \right) .
    \label{eq:DKLWithKilling}
\end{equation}
In the limit $n \to \infty, \Delta t_{\s} \to 0$ the sum over $\s$ is converted to an integral over time. Next, we choose
\begin{subequations}
    \begin{align}
        F_{\textrm{\H}}(x,t) &= b(x,t) + u(x,t) , \\
        F_{\textrm{\G}}(x,t) &= b(x,t) ,
    \end{align}
    \label{eq:DriftParameterization}
\end{subequations}
where $u(x,t)$ is called the control and we have the freedom to vary it, whereas $b(x,t)$ is fixed (uncontrolled). To extremize \cref{eq:DKLWithKilling} under the constraint \cref{eq:HPropagator}, we use the \textit{method of adjoint equation}, as explained in Sec.\ 2.3 of \cite{Chetrite:2021gcc}. In this approach, the optimal $u(x,t)$ and $P(x,t)$ are the stationary points of an action functional
\begin{equation}
    \begin{aligned}
        \mathcal{A}[\lambda,\P,u] =
        - \int_{-\infty}^{\infty} &\d x \left( \inP(x) \lambda(x,t_i) - \finP(x) \lambda(x,t_f) \right) \\
        &+ \int_{t_i}^{t_f} \d t \int_{-\infty}^{\infty} \d x \midP(x,t) \left( \frac{|u(x,t)|^2}{2 D(t)} + V_{\textrm{G}}(x,t) + (\pd_t + \textrm{L}_x) \lambda(x,t) \right) ,
        \label{eq:AdjointAction}
    \end{aligned}
\end{equation}
where $\L_x$ is the operator from \cref{eq:ForwardAndBackwardFP}, with $F = F_{\textrm{\H}}$. The bound from \cref{eq:DKLWithKilling} is clearly embedded in this action. The meaning of the other terms becomes apparent upon transferring the differential operators from $\lambda$ to $\midP$ through integration by parts,
\begin{equation}
    \mathcal{A}[\lambda,\P,u] = \int_{t_i}^{t_f} \d t \int_{-\infty}^{\infty} \d x \left( \frac{|u(x,t)|^2}{2 D(t)} + V_{\textrm{G}}(x,t) - \lambda(x,t) (\pd_t - \L_x^{\dagger}) \right) \midP(x,t) .
    \label{eq:Action}
\end{equation}
The field $\lambda(x,t)$ is a Lagrange multiplier which enforces the evolution law that the probability density $\midP(x,t)$ must obey, namely \cref{eq:HPropagator}. The action is stationary if the fields satisfy
%
\begin{subequations}
    \begin{align}
        &\pd_t \midP(x,t) = \L_x^{\dagger} \midP(x,t) , \label{eq:EOMofP} \\
        &(\pd_t + \L_x) \lambda(x,t) = -\frac{|u(x,t)|^2}{2 D(t)} - V_{\textrm{G}}(x,t) , \label{eq:DynamicProgramming} \\
        & \frac{u(x,t)}{D(t)} + \pd_x \lambda(x,t) = 0 \label{eq:ValueFunction} .
    \end{align}
    \label{eq:StationaryPoint}
\end{subequations}
The first of these is just the Fokker-Planck equation for $\midP$. We can combine \cref{eq:DynamicProgramming,eq:ValueFunction} to obtain the \textit{Hamilton-Jacobi-Bellman} (HJB) equation \cite{Pra1990}
\begin{equation}
    \pd_t \lambda(x,t) + b(x,t) \pd_x \lambda(x,t) + \frac{1}{2} D(t) \pd_x^2 \lambda(x,t) - D(t) \frac{|\pd_x \lambda(x,t)|^2}{2} = - V_{\textrm{G}}(x,t) .
\end{equation}
This can be mapped into a generalized backward Kolmogorov equation like \cref{eq:BackwardGeneralized} by the logarithmic transform
\begin{equation}
    \lambda(x,t) = -\ln \chi(x,t) . \label{eq:LogTransform}
\end{equation}
Then, of all the processes that transform $\inP \to \finP$ via \cref{eq:HPropagator}, the one which minimizes ${\rm D}_{KL}({\rm \H || \G})$ is specified by the pair of equations
\begin{subequations}
    \begin{align}
        \pd_t P(x,t) &= -\pd_x \left( [b(x,t) + D(t) \pd_x \ln \chi(x,t)] P(x,t) \right) + \frac{D(t)}{2} \pd_x^2 P(x,t) , \label{eq:OptimalFP} \\
        \pd_t \chi(x,t) &= -b(x,t) \pd_x \chi(x,t) - \frac{D(t)}{2} \pd_x^2 \chi(x,t) + V_{\textrm{G}}(x,t) \chi(x,t).
        \label{eq:ControlDE} 
    \end{align}
    \label{eq:OptimalControlDE}
\end{subequations}
The distribution $\midP(x,t)$ is called the \textit{Schr\"{o}dinger bridge} \cite{aebi1996}, and it  interpolates between $\inP$ and $\finP$, guided by the optimal control $u(x,t) = D(t) \pd_x \ln \chi(x,t)$. The kernel $\H^\star$ is the transition probability corresponding to \cref{eq:OptimalFP}. It can be shown by a direct calculation that
\begin{equation}
    H^\star(x, t | \xi, \tau) = \G(x, t | \xi, \tau) \frac{\chi(x,t)}{\chi(\xi,\tau)} .
    \label{eq:OptimalKernel}
\end{equation}
Since $G$ is fixed by the inherent dynamics of the system, it is $\chi$ that must adapt to the prescribed $\inP$ and $\finP$ via \cref{eq:HPropagator,eq:OptimalKernel}.
At the same time $\chi(x,t_f)$ determines $\chi(x,t_i)$ via \cref{eq:ControlDE}. Thus, the optimal control problem is solved if we can find two functions $\chi(x,t_i)$ and $\chi(x,t_f)$ which satisfy
\begin{subequations}
    \begin{align}
        \finP(x_f) &= \int_{-\infty}^{\infty} \d x_i \FTran{G}{f}{i} \frac{\chi(x_f, t_f)}{\chi(x_i, t_i)} \inP(x_i) \label{eq:OptimalPropagator} \\
        \chi(x_i, t_i) &= \int_{-\infty}^{\infty} \d x_f \, \chi(x_f, t_f) \FTran{G}{f}{i} \label{eq:OptimalEvolution} .
    \end{align}
    \label{eq:SchrodingerSystem}
\end{subequations}

\subsection{Reverse Diffusion}
\label{sec:ReversalOptimal}

It is difficult to find an analytic solution to \cref{eq:SchrodingerSystem} for a general $\inP, \finP$, and $\G$. However, such a solution is possible when $\inP$ is a diffused version of $\finP$. This is hardly surprising since it is well known that a diffusive process can be reversed by another diffusive process, of the same form as \cref{eq:OptimalFP} \cite{Nelson66,Anderson1982,Haussman86,Follmer88}. Nonetheless, casting reverse diffusion as an optimal control problem allows us to determine the action \cref{eq:AdjointAction} associated with this transformation.
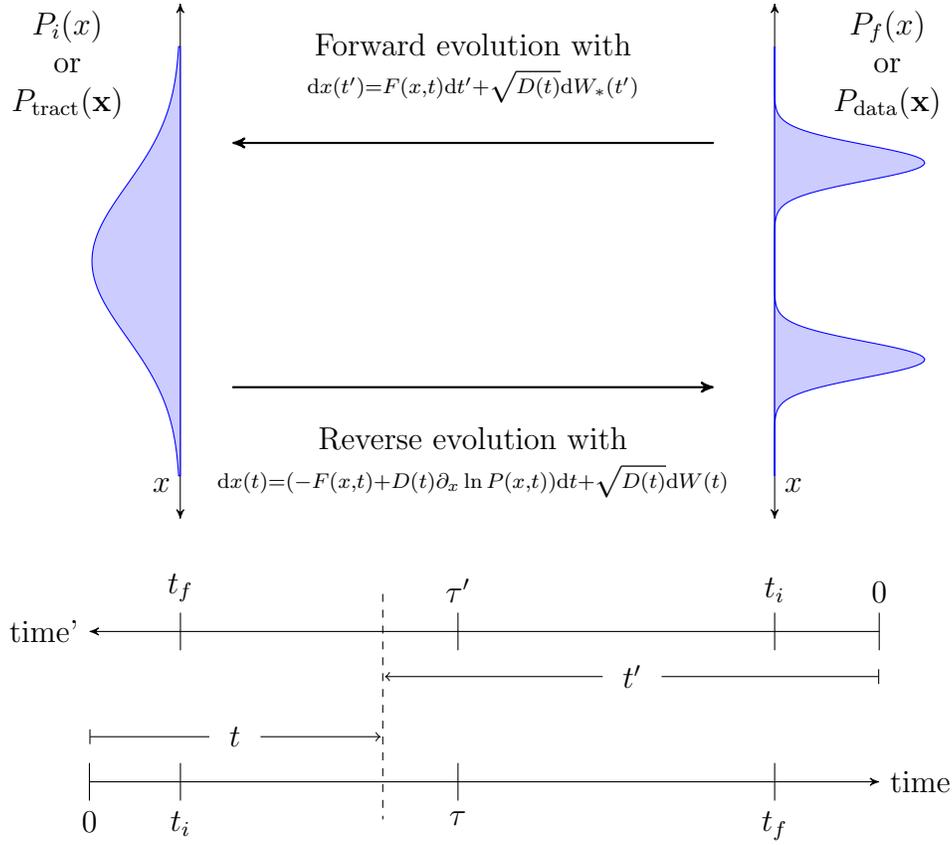
\begin{figure}
    \pgfmathdeclarefunction{gauss}{2}{%
      \pgfmathparse{1/(#2*sqrt(2*pi))*exp(-((x-#1)^2)/(2*#2^2))}%
    }
    
    \centering
    \begin{tikzpicture}
    \begin{scope}[shift={(-5,0)}]
        \begin{axis}[grid=none,
            ymax=2.1,
              axis lines=middle,
              y=1cm,
            y axis line style={draw=none},
            x axis line style={{stealth'}-{stealth'}},
            ytick=\empty,
            xtick=\empty,
            enlargelimits,
            rotate=90,
            ]
            \addplot[blue,fill=blue!20,domain=-2.4:2.4,samples=200]  {2.5*gauss(0,0.85)} \closedcycle;
    
            \node[align=center] at (axis cs:2.2,1.5) {$\inP(x)$ \\ or \\ $\w_{\rm tract}(\mathbf{x})$};
            \node at (axis cs:-2.5,0.25) {$x$};
        \end{axis}
    \end{scope}
    
    \draw[thick,-{stealth'}] (4.4,5) -- ++(-6.4,0) coordinate[pos=0.5] (a);
    \node[align=center,shift={(0,1)}] at (a) {Forward evolution with \\ $ \scriptstyle \d x(t') = F(x,t) \d t' + \sqrt{D(t)} \d W_*(t')$};
    
    \draw[thick,-{stealth'}] (-2,1.75) -- ++(6.4,0) coordinate[pos=0.5] (b);
    \node[align=center,shift={(0,-1)}] at (b) {Reverse evolution with \\ $ \scriptstyle \d x(t) = (-F(x,t) + D(t) \pd_x \ln \midP(x,t) ) \d t + \sqrt{D(t)} \d W(t)$};

    
    \begin{scope}[shift={(5,0)}]
        \begin{axis}[grid=none,
            ymax=2.1,
            axis lines=middle,
              y=1cm,
            y axis line style={draw=none},
            x axis line style={{stealth'}-{stealth'}},
            ytick=\empty,
            xtick=\empty,
            enlargelimits,
            rotate=-90,
            ]
            \addplot[blue,fill=blue!20,domain=-2.4:2.4,samples=200] {gauss(-1.1,0.2) + gauss(1.1,0.2)} \closedcycle;
    
            \node[align=center] at (axis cs:-2.2,1.5) {$\finP(x)$ \\ or \\ $\w_{\rm data}(\mathbf{x})$};
            \node at (axis cs:2.5,0.25) {$x$};
        \end{axis}
    \end{scope}

    \begin{scope}[shift={(0,-1.5)}]
        \draw[-{stealth'}] (6.6,0) -- (-3.9,0) node[left] {time'};
        \draw (6.6,0.25) node[above] {$0$} -- ++(0,-0.5);
        \draw (-2.69,0.25) node[above] {$t_f$} -- ++(0,-0.5);
        \draw (5.21,0.25) node[above] {$t_i$} -- ++(0,-0.5);
        \draw (1,0.25) node[above] {$\tau'$} -- ++(0,-0.5);
        \draw[|->] (6.6,-0.6) -- (0.03,-0.6) node[circle,fill=white,draw=none,pos=0.5] {$t'$};
    \end{scope}

    \draw[dashed] (0,-1) -- (0,-4);

    \begin{scope}[shift={(0,-3.5)}]
        \draw[-{stealth'}] (-3.9,0) -- (6.6,0) node[right] {time};
        \draw (-3.9,0.25) -- ++(0,-0.5) node[below] {$0$};
        \draw (-2.69,0.25) -- ++(0,-0.5) node[below] {$t_i$};
        \draw (5.21,0.25) -- ++(0,-0.5) node[below] {$t_f$};
        \draw (1,0.25) -- ++(0,-0.5) node[below] {$\tau$};
        \draw[|->] (-3.9,0.6) -- (-0.03,0.6) node[circle,fill=white,draw=none,pos=0.5] {$t$};
    \end{scope}
    
    
    \end{tikzpicture}
    \caption{\label{fig:ReverseDiffusion} A schematic representation of reverse diffusion. The time variables $t'$ and $t$ for the forward and reverse processes are also indicated. We set $t_i = 0$ and $t_f = T$ from \cref{sec:WorkingWithData} onward. The variables $\tau'$ and $\tau$ are used in \cref{sec:SpecialKernel}.}
 \end{figure}

The process of solving \cref{eq:SchrodingerSystem} is considerably simplified if we choose a killing rate
$V_{\textrm{G}}(x,t) = -\pd_x b(x,t)$, 
so that \cref{eq:ControlDE} becomes
\begin{equation}
    \pd_t \chi(x,t) = - \pd_x \left( b(x,t) \pd_x \chi(x,t) \right) - \frac{1}{2} D(t) \pd_x^2 \chi(x,t) .
    \label{eq:BackwardFokkerPlanck}
\end{equation}
This can be converted into a Fokker-Planck equation for $\arr{\chi}(x,t') \deq \chi(x,t)$, where $t' \deq t_f - (t-t_i)$ is a new time variable which runs from $t_i \to t_f$ as $t$ goes from $t_f \to t_i$ (see \cref{fig:ReverseDiffusion}):
\begin{equation}
    \pd_{t'} \arr{\chi}(x,t') = \pd_x \left( b(x,t_f + t_i - t') \pd_x \arr{\chi}(x,t') \right) + \frac{1}{2} D(t_f + t_i - t') \pd_x^2 \arr{\chi}(x,t') .
\end{equation}
The corresponding SDE is
\begin{align}
    \d x(t')
        &= - b(x, t_f + t_i - t') \d t' + \sqrt{D(t_f + t_i - t')} \d W_*(t') \nonumber \\
        &= -b(x,t) \d t' + \sqrt{D(t)} \d W_*(t') . \label{eq:ControlSDE}
\end{align}
where $W_*(t')$ has the same properties as $W(t)$, the Wiener term associated with $\textrm{\H}$ and $\textrm{\G}$ \cite{Nelson66}. In other words, \cref{eq:BackwardFokkerPlanck} is just a diffusion process in the direction of increasing $t'$, with drift $-b(x,t)$ and diffusion coefficient $D(t)$ (cf. \cref{eq:LangevinSDE}). If $\inP$ is the result of subjecting $\finP$ to a process
\begin{equation}
    \d x(t') =  F(x,t) \d t' + \sqrt{D(t)} \d W_*(t') , \label{eq:ForwardSDE}
\end{equation}
we can impose $\chi(x,t_f) = \finP(x)$ and $\chi(x,t_i) = \inP(x)$, by choosing
\begin{equation}
    b(x,t) = -F(x,t)  \label{eq:UncontrolledChoice}
\end{equation}
in \cref{eq:ControlSDE}. These boundary values of $\chi$ obey \cref{eq:SchrodingerSystem} (see \cref{sec:SchrodingerSystem}). The function $\arr{\chi}(x,t')$ that evolves from $\finP \to \inP$ is precisely the $\midP(x,t)$ that bridges $\inP \to \finP$, since $t$ and $t'$ label the same instant of time. Therefore, $\chi(x,t) = \arr{\chi}(x,t_f + t_i - t) = \midP(x,t)$, and the optimal control
%
%
\begin{equation}
    u = D(t) \pd_x \ln \midP(x,t) \label{eq:OptimalControlReversal}
\end{equation}
is proportional to the so called \textit{score function}, $\pd_x \ln \midP(x,t)$. 
Reverse diffusion is affected by the SDE
\begin{equation}
    \d x(t) = \left[ -F(x,t) + D(t) \pd_x \ln \midP(x,t) \right] \d t + \sqrt{D(t)} \d W(t) . \label{eq:ReversedLangevin}
\end{equation}
%
%
%
We made a specific choice of the process $\textrm{\G}$, with drift $-F(x,t)$ and killing rate $V_\textrm{\G}(x,t) = \pd_x F(x,t)$, to obtain this result. This is not just a mathematical trick; plugging these values in \cref{eq:AdjointAction} produces an action that resembles the cost functional used in \cite{Pavon89}, modulo terms involving the Lagrange multiplier. The cost functional is derived there from the \textit{Onsager-Machlup function} for continuous stochastic processes \cite{Yasue08}. The same functional appears in \cite{Huang21,berner2023optimal}, where it follows from the Feynman-Kac formula. We will now see how \cref{eq:AdjointAction} can be refashioned into a variational training objective for diffusion models.

\subsection{Score Matching from Least Action}
\label{sec:ScoreMatchingLeastAction}

The action in \cref{eq:AdjointAction} is stationary at a $\lambda, \midP$ and $u$ that satisfies \cref{eq:StationaryPoint}. Suppose we keep $\lambda$ and $\midP$ fixed at those values, but change $u$ to some other $\hat{u}$ consistent with the boundary conditions. The action changes by
%
%
\begin{align}
    \Delta \mathcal{A} &= 
    \mathcal{A}[\lambda,\P,\hat{u}] - \mathcal{A}[\lambda,\P,u] \nonumber \\
        &= \int_{t_i}^{t_f} \d t \int_{-\infty}^{\infty} \d x \, \midP(x,t)
        \bigg( \frac{|\hat{u}(x,t)|^2 - |u(x,t)|^2 }{2 D(t)} 
        + (\hat{u}(x,t) - u(x,t)) \pd_x \lambda(x,t) \bigg) \nonumber \\
        &= \int_{t_i}^{t_f} \d t \int_{-\infty}^{\infty} \d x \, \midP(x,t)
        \frac{\big| \hat{u}(x,t) - u(x,t) \big|^2}{2 D(t)} ,
\end{align}
where we have used \cref{eq:ValueFunction} in the second step to eliminate $\lambda$ in favor of $u$. This difference is positive definite, showing that $u$ is indeed a minima of $\mathcal{A}$. For reverse diffusion, the value of $u$ is given by \cref{eq:OptimalControlReversal}. If we parameterize $\hat{u} = D(t) S(x,t)$,
\begin{equation}
\boxed{
        \Delta \mathcal{A} =
            \frac{1}{2} \int_{t_i}^{t_f} \d t D(t) \int_{-\infty}^{\infty} \d x \, \midP(x,t)
            \big| S(x,t) - \pd_x \log \midP(x,t) \big|^2
        \label{eq:ChangeInAction}
    }
\end{equation}
This expression can also be understood as the pathwise Kullback-Leibler divergence between the process described by \cref{eq:ReversedLangevin}, and another one parameterized as\footnote{Use \cref{eq:ReversedLangevin,eq:LearnableProcess} (call these $\textrm{\H}$ and $\textrm{\G}$ respectively) in \cref{eq:DKLWithKilling}, with $V_{\textrm{\G}}$ set to zero.}
\begin{equation}
    \d x(t) = \left[ -F(x,t) + D(t) S(x,t) \right] \d t + \sqrt{D(t)} \d W(t) .
    \label{eq:LearnableProcess}
\end{equation}
%
We can convert $\inP \to \finP$ with \cref{eq:LearnableProcess} if we have a function $S(x,t)$ that minimizes \cref{eq:ChangeInAction}. This criterion can easily be translated into the score matching objective of \cite{Song2019,Song2021ScoreBased} for generative modeling applications.

\subsection{Working with Data}
\label{sec:WorkingWithData}

We have so far developed our intuition for the transforming probability distributions $\inP(x) \to \finP(x)$ by thinking of these distributions as the normalized density of a collection of diffusing particles. This mental picture is still useful when we are longer dealing with the position $x$ of a particle, but a data vector $\x$ in some higher dimensional space. We are given a large set of such initial data vectors, $\D = \{\x_{\rm d}\}$, that we presume is a faithful sampling of an underlying data distribution $\w_{\rm data}(\x)$. We do not have an analytic expression for this distribution, and our goal is to reconstruct $\w_{\rm data}(\x)$ from the data set $\D$. That is, we would like to generate new samples from $\w_{\rm data}(\x)$. The idea of using diffusion to accomplish this was first introduced in \cite{Sohl-DicksteinW15}.

If $\D$ is subjected to the dynamics in \cref{eq:ForwardSDE}, the data vectors are jostled around till their distribution approaches a form completely determined by drift and noise in the process. This is the forward stage of the algorithm. For specific choices of $\F$ and $\W$ the asymptotic distribution could take a simple, tractable form, which we designate as $\w_{\rm tract}(\x)$. We have already seen an example of this in \cref{sec:OUprocess}, where the final distribution is a Gaussian. Starting from $\w_{\rm tract}(\x)$ we can now reconstitute the original data distribution $\w_{\rm data}(\x)$ using the machinery developed in \cref{sec:ReversalOptimal}. This is the reverse stage. In practice, we pick a random vector from the tractable distribution $\w_{\rm tract}(\x)$ and subject it to the reverse dynamics till we arrive at a new sample of $\w_{\rm data}(\x)$.

Following \cref{sec:ReversalOptimal}, we use $t'$ to label time in the forward direction, and $t$ as the time variable for the reverse process. By our convention $\inP(x) \equiv \w_{\rm tract}(\x)$ and $\finP(x) \equiv \w_{\rm data}(\x)$, therefore reversal starts at time $t=t_i$ and ends at time $t=t_f$. The forward process runs from $t'=t_i$ to $t'=t_f$ (see \cref{fig:ReverseDiffusion}).
We can choose $t_i = 0$ and $t_f = T$ to abide by existing literature\footnote{To be precise, \cite{Sohl-DicksteinW15,Ho2020DDPM,Song2019} go from $0 \to T$ during the forward stage and from $T \to 0$ during the reverse stage, whereas we go from $0 \to T$ in both because we use two different time variables for each stage.} \cite{Sohl-DicksteinW15,Ho2020DDPM,Song2019,Huang21}, in which case $t' = T - t$.
Then, the forward evolution (cf.\ \cref{eq:ForwardSDE})
\begin{subequations}
    \begin{align}
        \d \x(t') &= \F(\x,t) \d t' + \sqrt{D(t)} \d \W_*(t')  \label{eq:LangevinData} \\
        \midP(\x,t) &= \int \d \x_{\rm d} \K(\x, t' | \x_{\rm d}, 0) \w_{\rm data}(\x_{\rm d}) , \label{eq:IntermediateData}
    \end{align}
\end{subequations}
is reversed by the SDE (cf.\ \cref{eq:ReversedLangevin})
%
\begin{equation}
    \d \x(t) = \underbrace{\left[ -\F(\x,t) + D(t) \nabla_\x \ln \midP(\x, t) \right]}_{\rdeq \F_{\rm rev} (\x,t)} \d t + \sqrt{D(t)} \d \W(t) .
    \label{eq:ReversedLangevinBackward}
\end{equation}
%
%
In \cref{eq:IntermediateData}, $\K$ denotes the forward transition probability associated with \cref{eq:LangevinData}. We do not have an analytic expression for $\midP(\x,t)$ however, since we do not know $\w_{\rm data}(\x_{\rm d})$. This issue can be circumvented in the end using a simple manipulation. We will assume for now that $\F_{\rm rev}(\x,t)$ can be numerically computed at all $\x$ that start at $\x_{\rm d}$ and are accessed by finite time evolution \cref{eq:LangevinData}. We can do this for each $\x_{\rm d} \in \D$, over many realizations of the stochastic trajectories. Thus, at the end of the forward pass $\F_{\rm rev}(\x,t)$ is known over a fine mesh of points in the ambient data space.
We try to match the reversal process with an SDE (cf.\ \cref{eq:LearnableProcess})
\begin{equation}
    \d \x(t) = [- \F(\x,t) + D(t) \S_{\th} (\x, t)] \d t + \sqrt{D(t)} \d \W(t) ,
    \label{eq:GenerativeSDE}
\end{equation}
where $\th$ are parameters of a neural network trained to minimize \cref{eq:ChangeInAction}. Replacing the square of real numbers with the standard $\ell^2$ norm for vectors in that expression, we have:\footnote{\cref{eq:PiecewiseDKLScore} is closely related to the theorems in \cite{Song2021Max}.}
\begin{equation}
    \Delta \mathcal{A} =
        \sum_{\s=1}^{n-1} \frac{1}{2} D(t_{\s}) \Delta t_{\s} \int \d \x_{\s} \midP(\x_{\s},t_{\s})  \left\lVert \S_{\th} (\x_{\s},t_{\s}) - \nabla_{\x_{\s}} \ln \midP(\x_{\s},t_{\s}) \right\rVert^2_2 .
    \label{eq:PiecewiseDKLScore}
\end{equation}
It is shown in \cite{Vincent2011} that the $\th^*$ which extremizes terms of this form can be learned from a \textit{denoising score matching} training objective that involves\footnote{In ML parlance, transition probability is all you need.} $\nabla_{\x_{\s}} \ln \K(\x_{\s}, t_{\s}|\x_{\rm d}, 0)$, rather than $\nabla_{\x_{\s}} \ln \midP(\x_{\s},t_{\s})$.
The discussion in \cref{sec:SpecialKernel} justifies this step, and some physical intuition for it is given in \cref{sec:MeaningOfScore}. Then,
%
\begin{equation}
    \th^* = \underset{\th}{\rm argmin}
        \sum_{\s=1}^{n-1} \frac{1}{2} D(t_{\s}) \Delta t_{\s} \,
        \mathbb{E}_{\x_{\rm d}} \mathbb{E}_{\x_{\s}|\x_{\rm d}}
        \bigg[ 
            \left\lVert \S_{\th} (\x_{\s},t_{\s}) - \nabla_{\x_{\s}} \ln \K(\x_{\s}, t_{\s} | \x_{\rm d}, 0) \right\rVert^2_2
        \bigg] .
    \label{eq:TrainingObjective}
\end{equation}
%
This is the fundamental training objective for generative diffusion models.
Importantly, it does not require that we know $\w_{\rm data}$; the expectation values are computed using the given data set $\D$, and our knowledge of how we numerically evolved each $\x_{\rm \d}$ to $\x_{\s}$ \cite{Hyvarinin2005}.
If there are $N$ data vectors in $\D$, and each $\x_{\rm d} \in \D$ arrives $M(\x_{\s}|\x_{\rm d})$ times at some $\x_{\s}$,
\begin{equation}
    \mathbb{E}_{\x_{\rm d}} \mathbb{E}_{\x_{\s}|\x_{\rm d}} [f(\x_{\s}; \x_{\rm d})]
        = \frac{1}{N}
            \sum_{\x_{\rm d} \in \D} \frac{1}{M(\x_{\s}|\x_{\rm d})}
            \sum_{\x_{\s}|\x_{\rm d}} f(\x_{\s}; \x_{\rm d}) .
    \label{eq:JointExpectation}
\end{equation}

Once the network is trained, it can interpolate the score function between points of the mesh that were explored during the forward pass. The quality of this interpolation depends on the neural network architecture, the original data set $\D$, and the granularity of the forward process \cite{Ho2020DDPM,Song2019,Song2021Max}. Starting with a random vector from $\w_{\rm tract}$ we can generate a new sample by evolving it with \cref{eq:GenerativeSDE}, this time using the learned $\S_{\th^*}$.
%
%
Further details of sampling, and refinements to it are given in \cite{Song2021ScoreBased,nichol2021improved}.

\section{Unifying Diffusion Models}
\label{sec:Unification}

The objective from \cref{eq:TrainingObjective} is the basis for several approaches to generative diffusion models. It appears explicitly in \textit{Score Matching with Langevin Dynamics} (SMLD) \cite{Song2019,Song2021ScoreBased} whereas it is implicit in \textit{Diffusion Probabilistic Modeling} (DPM) \cite{Sohl-DicksteinW15} and \textit{Denoising Diffusion Probabilistic Modeling} (DDPM) \cite{Ho2020DDPM}. SMLD uses a weighted combination of denoising score matching objectives, which reduces to \cref{eq:TrainingObjective} for a specific choice of the weighting function \cite{Song2021Max}. With a little work, DPM also fits into this framework.

\subsection{Diffusion Probabilistic Models}
\label{sec:SpecialKernel}

Suppose we subject a single particle to the forward process from \cref{eq:ForwardSDE}: we release the particle from $x_0$ at time $t_i$, wait a while, and find the particle at position $x$ at time $t'$ (see \cref{fig:ReverseDiffusion}). \textit{Given} these two pieces of information, the probability of locating the particle at $\xi$ at the intermediate time $\tau' \in (t_i, t')$ is
\begin{equation}
    \arr{\Q}(\xi, \tau' | x, t'; x_0, t_i) = \frac{\K(x, t' | \xi, \tau') \K(\xi, \tau' | x_0, t_i)}{\K(x, t' | x_0, t_i)} .
    \label{eq:BiconditionalProb}
\end{equation}
This expression follows from Bayes' theorem and the Markov property of the random process.\footnote{A more intuitive explanation of \cref{eq:BiconditionalProb} is given in \cite{Chetrite:2021gcc}.} The symbol $\K$ denotes probabilities associated with \cref{eq:ForwardSDE}, and are functions of the primed time variable.  We can use $\arr{\Q}$ to guess where the particle might have been at time $\tau' < t'$. The accuracy of our guess would improve if $\tau'$ is very close to $t'$, since the particle was most likely in the vicinity of its eventual destination $x$ at that time. That is, $\arr{\Q}$ becomes a sharply peaked function of $\xi$ centered around $x$, as $\tau'$ approaches $t'$. In fact, it is a Gaussian in this limit, as we show below.

The kernel $\arr{\Q}$ is fully determined from the transition probabilities for the forward stage. For an Ornstein-Uhlenbeck process these transition probabilities are Gaussian even for finite time, as seen in \cref{eq:OUkernel}. In that case $\arr{\Q}$ can be found by an explicit calculation and turns out to be a Gaussian as $\tau' \to t'$ \cite{Sohl-DicksteinW15,Ho2020DDPM}. This is generically true for any forward process described by \cref{eq:ForwardSDE}, under some mild assumptions. We can see this by differentiating \cref{eq:BiconditionalProb} with respect to $\tau'$ and substituting \cref{eq:ForwardAndBackwardFP}. With a little algebra, the result can be organized into
\begin{equation}
    \pd_{\tau'} \arr{\Q}(\xi, \tau' | x, t'; x_0, t_i) =
        - \pd_\xi ( F_1(\xi, \tau, x_0) \arr{\Q}(\xi, \tau' | x, t'; x_0, t_i) )
        - \frac{D(\tau)}{2} \pd_\xi^2 \arr{\Q}(\xi, \tau' | x, t'; x_0, t_i) \label{eq:ReverseEvolution1}
\end{equation}
%
where\footnote{The variables $\tau$ and $\tau'$ are analogous to $t$ and $t'$ from \cref{sec:WorkingWithData}, with $t' > \tau' \Leftrightarrow t < \tau$. See \cref{fig:ReverseDiffusion}.} $\tau = t_f + t_i - \tau'$, and the drift force $F_1$ is
\begin{equation}
    F_1(\xi, \tau, x_0) \deq F(\xi, \tau) - D(\tau) \pd_\xi \ln \K(\xi, \tau' | x_0, t_i) .
    \label{eq:ReversedDrift}
\end{equation}
We can rewrite \cref{eq:ReverseEvolution1} as a Fokker-Planck equation
%
\begin{equation}
    \pd_\tau \Q(\xi, \tau | x, t; x_0, t_f)
        = \pd_\xi \left( F_1(\xi, \tau, x_0) \Q(\xi, \tau | x, t; x_0, t_f) \right)
        + \frac{D(\tau)}{2} \pd_\xi^2 \Q(\xi, \tau | x, t; x_0, t_f) , \label{eq:ReverseEvolution2}
\end{equation}
for $\Q(\xi, \tau | x, t; x_0, t_f) \deq \arr{\Q}(\xi, \tau' | x, t'; x_0, t_i)$. Therefore $\Q$ is a Gaussian for $t \to \tau$ by \cref{eq:SmallTransitionWDrift}. $\Q$ is the probability of finding the particle at $(\xi,\tau)$ given that it was at $x$ at $t<\tau$ and will end up at $x_0$ at time $t_f$ in the reverse stage. Comparing \cref{eq:ReversedDrift,eq:ReverseEvolution2} with \cref{eq:ReversedLangevin}, it is apparent that the kernel $\Q$ reverse diffuses particles back to $x_0$.
%

If there are several particles distributed as $\finP(x_0)$ initially, their distribution at time $t$ is reversed by the kernel
\begin{equation}
    \R(\xi, \tau | x, t) = \frac{1}{\P(x,t)} \int \d x_0 \Q(\xi, \tau | x, t; x_0, t_f) \P(x, t; x_0, t_f) ,
    \label{eq:ReferenceKernelFromQ}
\end{equation}
%
%
where the joint probability $\P(x,t;x_0,t_f) \deq \K(x,t';x_0,t_i) =  \K(x, t' | x_0, t_i) \finP(x_0)$. We show in \cref{sec:ReverseKernel} that $\R$ does indeed reverse the forward process. We also observed in \cref{sec:ScoreMatchingLeastAction} that a trainable process like the one in \cref{eq:LearnableProcess}, with $S = S_\th$, can mimic $\R$ by minimizing
\begin{equation}
    \Delta \mathcal{A}
        = \textrm{D}_{KL} (\textrm{\R} || \textrm{\T})
        = \int \d \xi \d x \P(x,t) \R(\xi,\tau|x,t) 
            \ln \frac{\R(\xi,\tau|x,t)}{\T_\th(\xi,\tau|x,t)} ,
    \label{eq:DPMObjective1}
\end{equation}
where $\T_\th$ is the kernel for \cref{eq:LearnableProcess}. If we write $\T_\th = \int \d x \finP(x) \T_\th$, and use the continuum version of the log-sum inequality \cref{eq:LogSumInequality}, we can bound \cref{eq:DPMObjective1} as
\begin{equation}
    \Delta \mathcal{A} \leq
        \int \d \xi \d x \d x_0
        \P(x, t; x_0, t_f) \Q(\xi, \tau | x, t; x_0, t_f)
        \ln \frac{\Q(\xi, \tau | x, t; x_0, t_f)}{\T_\th(\xi,\tau|x,t)} + C ,
\end{equation}
where $C$ contains terms that do not involve $\T_\th$. We can take $t \to t_i, \tau \to t_f$, and split the first term on the right into a sum of integrals over many small intervals, just as we did in \cref{eq:DKLBound},\footnote{There could be edge effects as we take $\tau \to t_f$ in $\Q(\xi, \tau | x, t; x_0, t_f)$, which can be remedied as explained in \cite{Sohl-DicksteinW15}.}
\begin{align}
    \Delta \mathcal{A} &\leq
        \int \d x_{\s+1} \d x_{\s} \d x_0
        \K(x_s, t_s; x_0, t_i)
        \Q(x_{\s+1}| x_{\s}; x_0)
        \ln \frac{\Q(x_{\s+1} | x_{\s}; x_0)}{\T_\th(x_{\s+1} | x_{\s})} + C .
\end{align}
Since $\Q$ and $\T_\th$ are both Gaussian over small intervals, the bound simplifies to 
\begin{equation}
    \Delta \mathcal{A} \leq
        \sum_{s=0}^{n-1} \frac{1}{2} D(t_s) \Delta t_s \int \d x_{\s} \d x_0 \K(x_s, t_s; x_0, t_i)
        \big| S_\th(x,t) - \pd_x \ln \K(x_s,t_s|x_0,t_i) \big|^2 + C .
\end{equation}
Replacing the $x$'s with data vectors $\x$, we recover the training objective from \cref{eq:TrainingObjective}.

\subsection{Denoising Diffusion Probabilistic Models}
\label{sec:DDPM}

The DPM training objective from \cite{Sohl-DicksteinW15} was cast into a denoising score matching form for the first time in \cite{Ho2020DDPM}. \cref{eq:TrainingObjective} is already in that form, so it is instructive to compare it with the expression in \cite{Ho2020DDPM}. The data set $\D$ is subjected to an Ornstein-Uhlenbeck process in the forward stage, which ultimately distributes its contents according to $\N(0, \mathbf{1})$. The forward time evolution is a special instance of \cref{eq:OUSDE} with
\begin{subequations}
   \begin{align}
        D(t) &= \beta(t) , \\
        \beta_{t_{\s}} &\deq \beta(t_{\s}) \Delta t_{\s} .
    \end{align}
    \label{eq:SohlCondition}
\end{subequations}
%
%
Then, the transition probability for a small time interval, \cref{eq:SmallTransitionWDrift}, takes on the simple form
\begin{equation}
    \ITran{\K}{s+1}{s}
    = \N(\x_{\s+1}, \x_{\s} \sqrt{1 - \beta_{t_{\s}}}, \beta_{t_{\s}} \mathbf{1}) .
    \label{eq:ForwardSohl}
\end{equation}
where we have used the smallness of $\beta_{t_{\s}}$ to write $1- \frac{\beta_{t_{\s}}}{2} \approx \sqrt{1 - \beta_{t_{\s}}}$. Importing some more notation from \cite{Ho2020DDPM}, we define
\begin{align}
    \alpha_t &\deq 1 - \beta_t \approx e^{-\beta(t) \Delta t} , \\
    \ba{t} &\deq \prod_{s=0}^{n-1} \alpha_{t_s} \approx e^{- \int_{0}^{t} \beta(\bar{t}) \d \bar{t}} .
\end{align}
For the choice in \cref{eq:SohlCondition}, we can express the finite time transition probability, \cref{eq:OUkernel}, in terms of $\ba{t}$:
\begin{equation}
    \K(\x, t | \x_{\rm d}, 0) = \N(\x, \sqrt{\ba{t}} \x_{\rm d}, (1-\ba{t}) \mathbf{1}) .
    \label{eq:OUkernelHo}
\end{equation}
Samples drawn from this distribution can be written as $\x = \sqrt{\ba{t}} \x_{\rm d} + \sqrt{1-\ba{t}} \bm{\epsilon}$, where $\bm{\epsilon} \sim \N(0, \mathbf{1})$. Then,
\begin{equation}
    \nabla_{\x_{\s}} \ln \K(\x_{\s}, t_{\s} | \x_{\rm d}, 0) = - \frac{\bm{\epsilon}}{\sqrt{1-\ba{t_{\s}}}} .
\end{equation}
To match with the parameterization in \cite{Ho2020DDPM}, we will also write the score function as $\S_{\th} (\x,t) = -(1-\ba{t})^{-1/2} \bm{\epsilon}_\th (\x,t)$. Plugging these relations into \cref{eq:TrainingObjective}, we arrive at the DDPM training objective:
%
%
\begin{equation}
    \th^* = \underset{\th}{\rm argmin}
        \sum_{\s=1}^{n-1} 
        \frac{\beta_{t_{\s}}}{2 (1-\ba{t_{\s}})}
        \mathbb{E}_{\x_{\rm d}, \bm{\epsilon}}
        \bigg[ 
            \left\lVert \bm{\epsilon} - \bm{\epsilon}_\th(\sqrt{\ba{t_{\s}}} \x_{\rm d} + \sqrt{1-\ba{t_{\s}}} \bm{\epsilon}, t_{\s}) \right\rVert^2_2
        \bigg] .
    \label{eq:TrainingObjectiveHo}
\end{equation}
This is precisely the expression from \cite{Ho2020DDPM}, up to $O(\beta_t^2)$ corrections in the prefactor. Ultimately, \cite{Ho2020DDPM} uses a variant of \cref{eq:TrainingObjectiveHo} in their experiments.

\subsection{The Physical Meaning of Score}
\label{sec:MeaningOfScore}

{
\def \K {\midP}

We can use Brownian motion to build a \textit{rough} intuition for the reverse Langevin dynamics of \cref{eq:ReversedLangevinBackward} and the objective in \cref{eq:TrainingObjective}. We begin by considering a special case of \cref{eq:FokkerPlanck}, with a constant diffusion coefficient $D$, and a time-independent force $F(x)$ that is also confining in nature (see \cref{sec:OUprocess}). The noise $\eta$ tends to spread the particles around, broadening $\K(x, t)$ over time, whereas $F(x)$ tries to keep the particles localized to some region. Eventually, the particles settle into an equilibrium distribution $\K_{\rm eq}(x) \deq \K(x, t \to \infty)$, wherein the diffusion and confinement tendencies are evenly matched. Since $\K_{\rm eq}(x)$ is stationary, it must satisfy \cref{eq:FokkerPlanck}, with the time derivative set to zero. Then,
\begin{equation}
    S_{\rm eq}(x) = \frac{\d \ln \K_{\rm eq}(x)}{\d x} = \frac{2 F(x)}{D} . \label{eq:ScoreEqm}
\end{equation}
That is, the score of the equilibrium distribution is just the drift force acting on the particles, scaled by the strength of the noise. We can graft this intuition onto a nonequilibrium evolution under \cref{eq:FokkerPlanck}, provided it is sufficiently slow. 

Consider a simple diffusion process with no drift term, as given in \cref{eq:FPDiffusion}. This process has no equilibrium state and the particles will diffuse in perpetuity. The key idea is to imagine slow diffusion as a progression through equilibrium states. That is, we discretize the time variable to $t_s$ and map $\K(x,t_s)$ at each time slice to an equilibrium distribution, like the one from \cref{eq:ScoreEqm}. We may then think of $D(t_s) \pd_x \ln \K(x,t_s)$ as a force that was working against the noise to preserve the shape of $\K(x,t_s)$. This is all in our imagination of course; there is no such force in \cref{eq:FPDiffusion}. However, it does appear as the drift term in the reverse Langevin equation \cref{eq:ReversedLangevinBackward},
\begin{equation}
    \d x(t) = -D(t) \pd_x \ln \K(x, t) \d t + \sqrt{D(t)} \d W , \label{eq:ReverseDiffusionLangevin}
\end{equation}
where it confines and transforms $\K(x,t)$ at each time step till we recover the original distribution. Therefore, \cref{eq:ReverseDiffusionLangevin} takes us through the imaginary equilibrium states again but in the opposite direction. A similar intuition may be construed for a general Fokker-Planck process with $F(x,t) \neq 0$ using the fact that forces add vectorially.

\begin{figure}
    \pgfmathdeclarefunction{Power}{3}{%
      \pgfmathparse{#1*(x^2-#2^2)^2 + #3*x}%
    }
    \centering
    \begin{tikzpicture}
        \begin{axis}[grid=none,
              axis lines=middle,
            y axis line style={draw=none},
            xlabel={$x$},
            x label style={anchor=north},
            ytick=\empty,
            xtick=\empty,
            enlargelimits,
            legend style={at={(axis cs:1,1.25)}},
            xscale=1.25
            ]
        \addplot[blue,fill=blue!20,domain=-1.85:1.85,samples=200] {exp(-Power(1,1,0.25))} \closedcycle;
        \addplot[orange,thick,domain=-1.75:1.75,samples=200] {0.25*Power(1,1,0.25)};
        
        \legend{$\K_{\rm eq}(x)$, $V(x)$};
        

        \def \ss{\scriptstyle}
        \draw[thin,dashed] (axis cs:-1.03,0) node[anchor=south west] {$\ss x_1$} -- (axis cs:-1.03,1.29);
        \node[circle,fill=red,scale=0.25,draw=blue] at (axis cs:-1.1,1.26) {};
        \node[anchor=east] at (axis cs:-1.1,1.26) {$\ss A_1$};
        \node[circle,fill=red,scale=0.25,draw=blue] at (axis cs:-0.96,1.26) {};
        \node[anchor=west] at (axis cs:-0.96,1.26) {$\ss B_1$};

        \draw[thin,dashed] (axis cs:-0.5,0) node[anchor=south west] {$ \ss x_2$} -- (axis cs:-0.5,0.65);
        \node[circle,fill=red,scale=0.25,draw=blue] at (axis cs:-0.54,0.69) {};
        \node[anchor=east] at (axis cs:-0.54,0.69) {$\ss A_2$};
        \node[circle,fill=red,scale=0.25,draw=blue] at (axis cs:-0.43,0.57) {};
        \node[anchor=west] at (axis cs:-0.43,0.57) {$\ss B_2$};
        
        \end{axis}
    \end{tikzpicture}
    \caption{\label{fig:ScoreAsForce} The probability distribution for a process in which the drift and diffusion tendencies are in equilibrium (cf.\ \cref{eq:EquilibriumProb}). Particles are concentrated at the minima of the potential $V(x)$ and stay away from its maxima. Perturbing this distribution gently allows us to estimate the score.}
\end{figure}
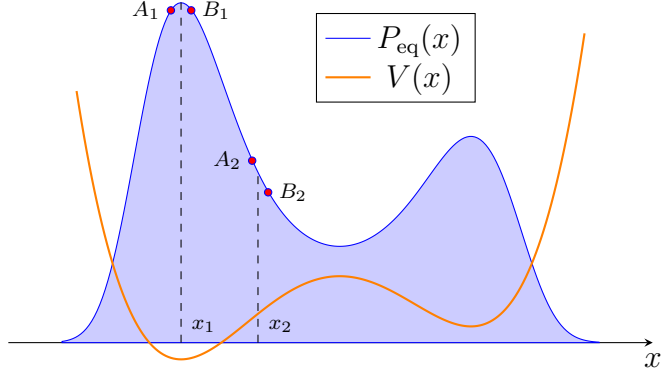

The idea of the score function $S(x,t_s)$ as a force field is also helpful in understanding the denoising score matching objective in \cref{eq:TrainingObjective}. We can solve \cref{eq:ScoreEqm} by writing the force as $F(x) = -\d V(x) / \d x$, where $V(x)$ is a potential energy. Then,
\begin{equation}
    \K_{\rm eq}(x) = \frac{1}{Z} \exp(-\frac{2 V(x)}{D}). \label{eq:EquilibriumProb}
\end{equation}
In particular, the maximas of $\K_{\rm eq}(x)$ correspond to the minimas of $V(x)$, and vice versa. This makes physical sense: particles will flock to regions of lower energy, and stay away from areas with higher energy (see \cref{fig:ScoreAsForce}). Since force is the negative gradient of the potential, the force vector at each point is directed toward regions of lower energy in its immediate vicinity. This intuition also applies to data vectors evolving under \cref{eq:LangevinData}.


Suppose we were given a data set $\D$ that faithfully samples $\K_{\rm eq}(\x)$. The data points in $\D$ form a dense cloud with the same shape as $\K_{\rm eq}(\x)$ in a $d$-dimensional vector space. If we add a small amount of random white noise to each element in $\D$, the points will be jostled around a little, and every point in space will see a tiny influx and/or outflow of noised data. We repeat this experiment several times. That is,
\begin{enumerate}
    \item \label{itm:Perturbation} For each $\x$ in $\D$ compute $\tilde \x = \x + \bm{\eta}$, where $\bm{\eta}\sim \N(0, \sigma^2 \mathbf{1})$, and $\sigma$ is very small. Store the results as $\tilde \D_1$.
    \item Repeat the first step $M (\gg 1)$ times, storing the results in $\tilde \D_1, \tilde \D_2, \dots, \tilde \D_M$.
    \item Focus on a small hypercube $\d^d \x$ centered at each $\x$, averaging over all $\tilde \D_m$ to find the net flux through it. Do this for hypercubes over the whole vector space.
\end{enumerate}
The flow at each $\x$ tells us something about the potential $V(\x)$, and the force $\F(\x) = -\nabla_\x V(\x)$. First, the net flux through the maximas and minimas of $\K_{\rm eq}(\x)$ will be zero; there is a symmetric distribution of data points around such extrema, and flows from diametrically opposite points around each extrema cancel\footnote{This is true only if the noise itself is isotropic, which is why we add white noise and average over $M$ iterations.} (see for e.g.\ the points $A_1$ and $B_1$ around the peak at $x_1$ in \cref{fig:ScoreAsForce}). Since the maximas and minimas of $\K_{\rm eq}(\x)$ correspond to minimas and maximas of $V(\x)$, the force is zero at those points. Similarly, if there is a non-zero average flux in some particular direction through $\x$, like $A_2 \to B_2$ in \cref{fig:ScoreAsForce}, it means there are more data points at $A_2$ than at $B_2$. Therefore $A_2$ has a lower potential energy than $B_2$, and the force points from $B_2 \to A_2$. It must be possible then to reconstruct the force field $\F(\x)$, and ultimately the score function $\S(\x)$, from the fluxes (cf.\ \cref{eq:ScoreEqm}).

We can extend this intuition to all the intermediate distributions $\K(\x,t_s)$ of a slow diffusion process. In fact, \cref{eq:TrainingObjective} codifies the procedure we outlined above. For simplicity, we take the diffusion coefficient to be a constant and use \cref{eq:PureDiffusionKernel} to find
\begin{equation}
    \th^* = \underset{\th}{\rm argmin}
        \sum_{\s=1}^{n-1} \frac{1}{2} D \Delta t_{\s} \,
        \mathbb{E}_{\x_{\s}, \x_{\rm d}}
        \bigg[ 
            \left\lVert \S_{\th} (\x_{\s},t_{\s}) + \frac{\x_\s - \x_{\rm d}}{D t_\s} \right\rVert^2_2
        \bigg] .
\end{equation}
%
For small values of $t_{\s}$ the perturbation is minuscule, as in our thought experiment.
The expectation over $\x_{\rm d}$ adds up all the noise vectors that land on $\x_{\s}$, producing a vector commensurate with the score (or force) at that point. For larger $t_s$, the distribution $\K(\x,t_s)$ is the result of adding a small noise with $\sigma^2 = D \Delta t_s$ to $\K(\x_{\s-1}, t_{\s-1})$ (cf.\ \cref{eq:Trotterized}). Therefore, the score computed at that $t_s$ is proportional to the force on $\K(\x_{\s-1}, t_{\s-1})$.

} 


\section{Concluding Remarks}

We have shown that score matching can be derived from an action principle, by viewing reverse diffusion in terms of stochastic optimal control. This perspective allows us to understand score matching as an optimization problem, of finding the most efficient paths that transform a noisy vector back to a sample representing the training data. 
One hopes that the action principle generalizes in a straightforward manner to nontrivial data manifolds so that underlying symmetries of the data distribution can be modeled more accurately \cite{Bortoli22,jagvaral2022modeling}. This will be the subject of future work. 

\paragraph{Acknowledgements:} We are grateful to Evgueni Alexeev, Brian Campbell-Deem, and Hayden Lee for comments on the draft, to Austin Joyce for valuable discussions, and to the authors of \cite{Chetrite:2021gcc} for the English translation of Schr\"{o}dinger's paper \cite{schrodinger1931}.

\appendix
\section{Appendix}

\subsection{The Kullback-Leibler Divergence}
\label{sec:KLDivergence}


We review some properties of the Kullback-Leibler divergence, ${\rm D}_{KL}$, that were used in the main text. First, we use the inequality $\ln(1/x) \geq 1-x$ to write \cref{eq:DKLDiscrete} as
\begin{equation}
    {\rm D}_{KL}({\rm h||g}) \geq \sum_{kl} p_i(l) \h(k|l) \left( 1 - \frac{\h(k|l)}{\g(k|l)} \right) = 0.
    \label{eq:DKLPositivity}
\end{equation}
Next, we use the Markov property of the processes $\textrm{H}$ and $\textrm{G}$ to rewrite \cref{eq:DKLContinuum} as
\begin{align}
    &{\rm D}_{KL}({\rm \H || \G}) \nonumber \\
        &= \int \d x_f \d x \d x_0 \midP(x_i,t_i)
            \FTran{\H}{f}{}
            \FTran{\H}{}{i}
            \ln \frac{
                \int \d x \FTran{\H}{f}{} \FTran{\H}{}{i}
                }{
                    \int \d x \FTran{\G}{f}{} \FTran{\G}{}{i}
                } \nonumber \\[0.5em]
        &\leq \int \d x_f \d x \d x_i \midP(x_i,t_i)
            \FTran{\H}{f}{} \FTran{\H}{}{i}
            \ln \frac{\FTran{\H}{f}{} \FTran{\H}{}{i}}{\FTran{\G}{f}{} \FTran{\G}{}{i}} .
\end{align}
In the last step, we have used a continuum version of the log sum inequality,
\begin{equation}
     \left( \sum_{i=1}^{n} a_i \right) \log \left( \frac{\sum_{i=1}^{n} a_i}{\sum_{i=1}^{n} b_i} \right)
     \leq \sum_{i=1}^{n} a_i \log \frac{a_i}{b_i} , \,\text{ for } a_i, b_i \geq 0 , \label{eq:LogSumInequality}
\end{equation}
to remove the integrals inside the $\log$. Expanding this log and noting that $H$ can propagate $\midP(x_i,t_i)$ to the probability at any intermediate time $u$,
\begin{align}
    {\rm D}_{KL}({\rm \H || \G}) \leq
        \int \d x_f \d x \w(x, t)
            &\FTran{\H}{f}{} \ln \frac{\FTran{\H}{f}{}}{\FTran{\G}{f}{}} \\
        &+ \int \d x \d x_0 \midP(x_i,t_i)
            \FTran{\H}{}{i}
            \ln \frac{\FTran{\H}{}{i}}{\FTran{\G}{}{i}} . \nonumber
\end{align}
We can repeat this process on each term in the r.h.s.\ till we obtain $n+1$ partitions of the original interval $[t_i, t_f]$. With $t_i = t_0$ and $t_f = t_n$, and denoting $\FTran{\H}{\s+1}{\s} \equiv \Tran{\H}{\s+1}{\s}$ etc. for brevity,
\begin{equation}
    {\rm D}_{KL}({\rm \H || \G})
        \leq \sum_{\s=0}^{n-1} \int \d x_{\s+1} \d x_{\s} \, \midP(x_{\s},t_{\s}) \Tran{\H}{\s+1}{\s} \ln \frac{\Tran{\H}{\s+1}{\s}}{\Tran{\G}{\s+1}{\s}} .
    \label{eq:PathwiseKL}
\end{equation}
In the $n \to \infty$ limit the r.h.s., if it exists, becomes the \textit{pathwise} Kullback-Leibler divergence. The advantage of working with the pathwise divergence is that, for Fokker-Planck dynamics, the transition probabilities in an infinitesimal time interval have the simple form in \cref{eq:SmallTransitionWDrift}.

\subsection{Solution of the Schr\"{o}dinger system}
\label{sec:SchrodingerSystem}

We want to show that a function $\chi(x,t)$ that solves
\begin{subequations}
    \begin{gather}
        \pd_t \chi(x,t) = \pd_x \left( F(x,t) \pd_x \chi(x,t) \right) - \frac{1}{2} D(t) \pd_x^2 \chi(x,t) \label{eq:ControlDE2} \\
        \chi(x, t_f) = \finP(x) \label{eq:ControlBoundary1} \\
        \chi(x, t_i) = \inP(x) , \label{eq:ControlBoundary2}
    \end{gather}
    \label{eq:ControlDEReversal}
\end{subequations}
also satisfies \cref{eq:SchrodingerSystem} if $\finP$ is evolved to $\inP$ under \cref{eq:ForwardSDE}. Comparing \cref{eq:ControlDE2} with \cref{eq:BackwardGeneralized}, we see that any boundary conditions on $\chi$ must be related via \cref{eq:OptimalEvolution},
\begin{equation}
    \chi(x_i, t_i) = \int_{-\infty}^{\infty} \d x_f \, \chi(x_f, t_f) \FTran{G}{f}{i} ,
    \label{eq:OptimalEvolution2}
\end{equation}
where $G$ is the kernel corresponding to a killed diffusion process with the killing rate $V_{\textrm{\G}}(x,t) = \pd_x F(x,t)$ (cf.\ \cref{eq:FeynmanKac}). To see whether \cref{eq:ControlBoundary1,eq:ControlBoundary2} obey \cref{eq:OptimalEvolution2} we rewrite \cref{eq:ControlDE2} in terms of the function $\arr{\chi}(x,t') = \chi(x,t)$ defined in \cref{sec:ReversalOptimal},
%
%
\begin{equation}
        \pd_t \arr{\chi}(x,t') = -\pd_x \left( F(x,t) \pd_x \arr{\chi}(x,t') \right) + \frac{1}{2} D(t) \pd_x^2 \arr{\chi}(x,t') \label{eq:BackwardFP2}
\end{equation}
If $\arr{G}$ is the kernel corresponding to \cref{eq:BackwardFP2}, we have
\begin{subequations}
    \begin{align}
        \arr{\chi}(x_i, t'=t_f) &= \int_{-\infty}^{\infty} \d x_f \, \arr{G}(x_i, t'=t_f | x_f, t'=t_i) \arr{\chi}(x_f, t'=t_i) \\ \implies
        \chi(x_i, t_i) &= \int_{-\infty}^{\infty} \d x_f \, \arr{G}(x_i, t'=t_f | x_f, t'=t_i) \chi(x_f, t_f) \label{eq:OptimalEvolution3}
    \end{align}
\end{subequations}
%
It follows from \cref{eq:OptimalEvolution2,eq:OptimalEvolution3} that\footnote{The identification of $\G$ with $\arr{\G}$ here is equivalent to the approach in Sec.\ 3 of \cite{Huang21}.}
\begin{equation}
    \FTran{G}{f}{i} \equiv \arr{G}(x_i, t'=t_f | x_f, t'=t_i) . \label{eq:FKTrick}
\end{equation}
Furthermore, \cref{eq:BackwardFP2} corresponds to the process \cref{eq:ForwardSDE}, so that
\begin{equation}
    \inP(x_i) = \int \d x_f \, \arr{G}(x_i, t'=t_f | x_f, t'=t_i) \finP(x_f) .
\end{equation}
Therefore, \cref{eq:OptimalEvolution2} is indeed satisfied by \cref{eq:ControlBoundary1,eq:ControlBoundary2}. Finally, these boundary values transform \cref{eq:OptimalPropagator} into
\begin{equation}
    1 = \int \d x_i \FTran{G}{f}{i} ,
\end{equation}
which is true by \cref{eq:FKTrick}, and the fact that $\arr{\G}$ is a normalized transition probability. $\blacksquare$

\subsection{The DPM Kernel}
\label{sec:ReverseKernel}

We want to verify that the kernel from \cref{eq:ReferenceKernelFromQ} reverses the diffusive process in \cref{eq:ForwardSDE}. Suppose we use this kernel to propagate a distribution $\midP(x,t)$ to
\begin{equation}
    \midP(\xi,\tau) = \int \d x \, \R(\xi, \tau | x, t) \midP(x,t) .
    \label{eq:ReversePropagation1}
\end{equation}
Using \cref{eq:ReverseEvolution2}, it is straightforward to see that
\begin{align}
    \pd_\tau \midP(\xi,\tau) &=
        \pd_\xi (F(\xi,\tau) \midP(\xi,\tau)) + \frac{D(\tau)}{2} \pd_\xi^2 \midP(\xi,\tau) \label{eq:ReversePropagation2} \\
        &\quad
        - D(\tau) \pd_\xi
        \left(
            \int \d x_0 \d x \, \pd_\xi \ln \K(\xi,\tau'|x_0,t_i) \Q(\xi, \tau | x, t; x_0, t_f) \P(x, t; x_0, t_f)
        \right) . \nonumber
\end{align}
The last term simplifies to
\begin{equation}
    \begin{aligned}
        \int \d x_0 \, \pd_\xi \ln &\K(\xi,\tau'|x_0,t_i) \int \d x \Q(\xi, \tau | x, t; x_0, t_f) \P(x, t; x_0, t_f) \\
        &= \int \d x_0 \, \pd_\xi \ln \K(\xi,\tau'|x_0,t_i) \K(\xi,\tau';x_0,t_i) \\
        &= \pd_\xi \int \d x_0 \, \K(\xi,\tau'|x_0,t_i) \finP(x_0) \\
        &= \K(\xi,\tau') \pd_\xi \ln \K(\xi,\tau') .
    \end{aligned}
\end{equation}
It also follows from \cref{eq:ReversePropagation1}, and the definitions of $\Q$ and the joint probability (see \cref{sec:SpecialKernel}), that
\begin{equation}
    \begin{aligned}
        \midP(\xi,\tau)
            &= \int \d x_0 \d x \, \arr{\Q}(\xi, \tau' | x, t'; x_0, t_i) \K(x,t';x_0,t_i) \\
            &= \int \d x_0 \K(\xi,\tau';x_0,t_i) \\
            &= \K(\xi, \tau') .
    \end{aligned}
\end{equation}
Therefore, $\midP(\xi,\tau)$ evolves as
\begin{align}
    \pd_\tau \midP(\xi,\tau) &=
        \pd_\xi (\left[ F(\xi,\tau) - D(\tau) \pd_\xi \ln \midP(\xi,\tau) \right] \midP(\xi,\tau)) + \frac{D(\tau)}{2} \pd_\xi^2 \midP(\xi,\tau) .
\end{align}
$\blacksquare$

\clearpage
\phantomsection
\addcontentsline{toc}{section}{References}
\small
\bibliographystyle{utphys}
\bibliography{Diffusion}

\end{document}